\documentclass[journal]{IEEEtran}
\usepackage{amsmath,amsfonts}
\usepackage{algorithmic}
\usepackage{algorithm}
\usepackage{array}
\usepackage[caption=false,font=normalsize,labelfont=sf,textfont=sf]{subfig}
\usepackage{textcomp}
\usepackage{stfloats}
\usepackage{url}
\usepackage{verbatim}
\usepackage{graphicx}
\usepackage{cite}
\usepackage{amsmath,amsfonts}
\usepackage{algorithmic}
\usepackage{algorithm}
\usepackage{array}
\usepackage[caption=false,font=normalsize,labelfont=sf,textfont=sf]{subfig}
\usepackage{textcomp}
\usepackage{stfloats}
\usepackage{url}
\usepackage{verbatim}
\usepackage{graphicx}
\usepackage{cite}
\usepackage{makecell}
\usepackage[T1]{fontenc}
\begin{document}
\title{DOEPatch: Dynamically Optimized Ensemble Model for Adversarial Patches Generation}

\author{Wenyi Tan, Yang Li, Chenxing Zhao, Zhunga Liu, and Quan Pan
\thanks{\textit{(Corresponding author: Yang Li.)}}
\thanks{Wenyi Tan, Yang Li, Chenxing Zhao, Shuangju Zhou, Zhunga Liu, and Quan Pan are with School of Automation, Northwestern Polytechnical University, Xi'an 710072, China  (e-mail: tanwenyi@mail.nwpu.edu.cn, liyangnpu@nwpu.edu.cn, zhaochenxing@mail.nwpu.edu.cn).}
}
    
\markboth{IEEE TRANSACTIONS ON INFORMATION FORENSICS AND SECURITY,~Vol.~14, No.~8, August~2023}%
{Shell \MakeLowercase{\textit{et al.}}: A Sample Article Using IEEEtran.cls for IEEE Journals}



\maketitle

\begin{abstract}
Object detection is a fundamental task in various applications ranging from autonomous driving to intelligent security systems. However,  recognition of a person can be hindered when their clothing is decorated with carefully designed graffiti patterns, leading to the failure of object detection. To achieve greater attack potential against unknown black-box models, adversarial patches capable of affecting the outputs of multiple-object detection models are required. While ensemble models have proven effective, current research in the field of object detection typically focuses on the simple fusion of the outputs of all models, with limited attention being given to developing general adversarial patches that can function effectively in the physical world. In this paper, we introduce the concept of energy and treat the adversarial patches generation process as an optimization of the adversarial patches to minimize the total energy of the ``person'' category. Additionally, by adopting adversarial training, we construct a dynamically optimized ensemble model. During training, the weight parameters of the attacked target models are adjusted to find the balance point at which the generated adversarial patches can effectively attack all target models.
We carried out six sets of comparative experiments and tested our algorithm on five mainstream object detection models. The adversarial patches generated by our algorithm can reduce the recognition accuracy of YOLOv2 and YOLOv3 to 13.19\% and 29.20\%, respectively.
In addition, we conducted experiments to test the effectiveness of T-shirts covered with our adversarial patches in the physical world and could achieve that people are not recognized by the object detection model. Finally, leveraging the Grad-CAM tool, we explored the attack mechanism of adversarial patches from an energetic perspective.
\end{abstract}


\section{Introduction}
\label{sec:1}
 \IEEEPARstart{T}{he} rapid development of artificial intelligence has raised major security concerns in several fields, including image processing\cite{akhtar2018threat}, reinforcement learning\cite{LI2023103259,li2022deep}, natural language processing~\cite{li2021graph}, and unmanned aerial systems\cite{wang2023survey,wu2023highly}. Object detection models in deep learning, in particular, are vulnerable to adversarial samples, which can introduce carefully crafted perturbation\cite{goodfellow2014explaining,papernot2016limitations,szegedy2013intriguing}, patch\cite{brown2017adversarial,lee2019physical}, or camouflage\cite{wang2022fca,suryanto2022dta} into digital images and cause wrong predictions. This has emerged as a significant threat to the security of object detection models. Recently, a large amount of research has shown that adversarial examples are transferable\cite{moosavi2017universal,papernot2017practical,lin2019nesterov}, meaning that adversarial examples designed for one model can mislead other black-box models. Specifically, Cai et al.\cite{cai2023ensemble} proposed an adversarial attack method against dense prediction models. The adversarial examples generated by this method can successfully attack multiple object detection and segmentation models. 
Simultaneously, with the emergence of adversarial patches, recent research has proposed various attack methods based on these patches, which can be directly printed on cardboard\cite{thys2019fooling} and clothing\cite{wu2020making}, thereby impacting the recognition of object detection models in the physical world. This poses significant threats to areas such as pedestrian monitoring and autonomous driving. 

\begin{figure}[!t]
\centering
\includegraphics[scale=0.2]{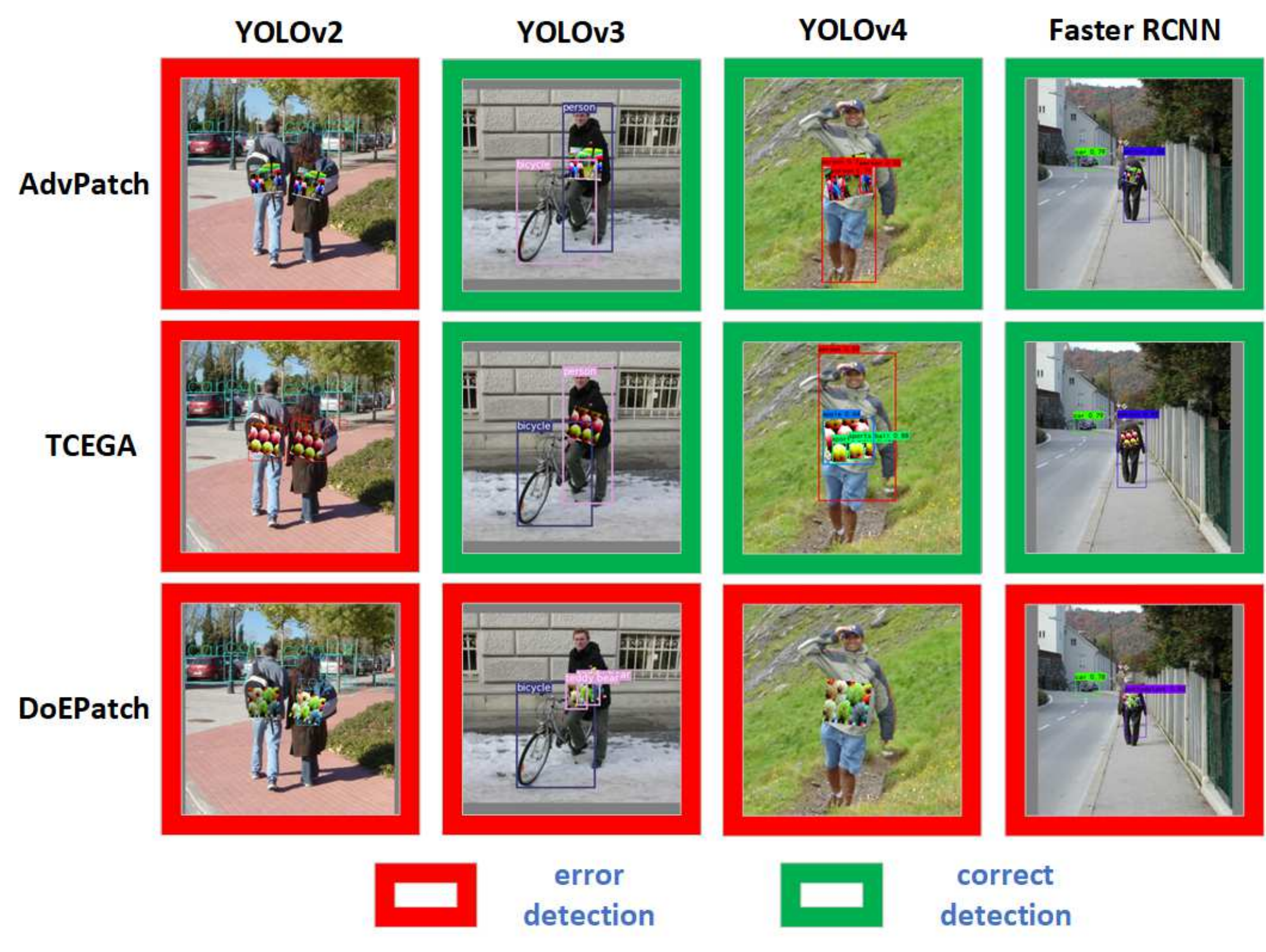}
\caption{The detection results of mainstream object detection models, on the adversarial patches, namely AdvPatch, TCEGA, and DOEPatch. The red bounding box denotes mistaken detection or the failure of the object detector to recognize the human, whereas the green bounding box signifies correct detection. }
\label{fig:1}
\end{figure}

In practical application scenarios, adversarial patches often confront unknown object detection models. However, the existing studies on adversarial patches have mainly emphasized the deployment of patches in the physical world\cite{xu2020adversarial} and attacks under multiple viewpoints\cite{hu2022adversarial}, the problem of the transferability of adversarial patches has been largely ignored. We generated several sets of adversarial samples randomly from the testing dataset. 
Apparently, the adversarial patches designed specifically for the YOLOv2\cite{redmon2017yolo9000} object detection model, such as AdvPatch\cite{thys2019fooling} and TCEGA\cite{hu2022adversarial}, are difficult to affect the detection results of other models such as YOLOv3\cite{redmon2018yolov3}, YOLOv4\cite{bochkovskiy2020yolov4}, and Faster R-CNN\cite{ren2015faster}(Fig.\ref{fig:1}). And our generated Dynamically Optimized Ensemble Patch(DOEPatch) was capable of incapacitating all the object detection models.

The various network architectures used in different object detection models exhibit diversity, and the feature maps obtained for the same sample are also distinct. Adversarial patches mainly disrupt the detection of the original target by the object detection model through special feature distributions generated during training. Therefore, adversarial patches tailored to a single object detection model are difficult to interfere with the detection of other unknown models. 
Although the transferability issue of adversarial samples has been widely studied, and ensemble-based methods have been widely used with significant effects in addressing the transferability problem.   However, earlier ensemble models\cite{liang2022large,huang2021rpattack} simply combined the outputs of different models without specifically balancing the weight parameters between victim models, which easily led to the optimization of adversarial patches in a singular direction, limiting the attack performance of ensemble models.

To improve the transferability of adversarial patches using ensemble models and explore the attack mechanisms for generating such patches,  we conduct an energy-based analysis to gain insights into the generation process. We then develop a universal adversarial patches attack framework by combining this analysis with a dynamic optimization approach for an ensemble model. Specifically, the energy-based model\cite{lecun2006tutorial,grathwohl2019your} generates low-energy samples from random noise inputs and subsequently enhances the quality of generated samples by updating the network parameters. We introduce the concept of energy-based models to reconstruct the energy function for generating adversarial patches targeted toward object detection models. This is achieved by reducing the energy of the ``person'' category to effectively launch an attack on the object detectors. During the training of adversarial patches, the confidence scores of bounding boxes related to the ``person'' category are incorporated into the energy-based function.  
Through iterative optimization, the energy of the generated adversarial patches is gradually increased, while the energy associated with the target category is reduced. 

To enhance the generalization of adversarial patches and strengthen their transferability across different object detection models, we utilize the concept of Min-Max adversarial training to construct a dynamically optimized ensemble model.  
Therefore, the process of adversarial patch generation can be divided into two main stages. The first stage entails internal maximization, wherein the weights of the target object detection model are dynamically adjusted to maximize the overall energy output. The second stage involves external minimization, wherein the adversarial patch is updated to minimize the overall energy output. Finally, by continuously undergoing Min-Max adversarial training and dynamically adjusting the weight parameters of the models, an adversarial patch that is effective against all target object detection models can be generated and achieved a balanced outcome. 

In this study, we developed six types of adversarial patches tailored to mainstream one-stage and two-stage object detection models. We conducted a thorough evaluation of the attack effectiveness and transferability of these patches across five object detection models. 
Our experimental results demonstrate the transferability of the proposed adversarial patches. 
In addition, we have achieved the ability to influence the recognition of a person by object detection models in the physical world by printing adversarial patches on clothing. 
Finally, we performed an explanatory analysis of the effects of our adversarial patches on object detection predictions by examining the variations in target category energy and visualizing feature distributions using Grad-CAM\cite{selvaraju2017grad} on adversarial samples. The main contributions of this work can be summarized as follows: 
\begin{itemize}
    \item We proposed novel universal adversarial patches, based on an energy function and a dynamically optimized ensemble model, that can be applied to multiple object detection models.  
    \item The effectiveness and transferability of the proposed adversarial patches were successfully demonstrated on various state-of-the-art object detection models, including physical-world attacks by printing the patches on clothing. 
    \item We conducted an explanatory analysis of the adversarial patches’ attacks from an energy perspective and visualized feature distributions using Grad-CAM. 
\end{itemize}

The sections are organized as followed: Section~\ref{sec:2} summarizes the related work on ensemble models and adversarial patches. Section~\ref{sec:3} describes the proposed algorithm. Section~\ref{sec:4} presents the experimental setup, corresponding results, and analysis. Section~\ref{sec:5} provides the conclusion for this paper. 


\section{Related Works}
\label{sec:2}
This paper mainly studies the adversarial patch attack method based on the ensemble model, aiming to generate a universal adversarial patch that can attack multiple object detection models simultaneously. Therefore, we mainly introduce the work about ensemble-based attacks and adversarial patch attacks. 
\subsection{Ensemble-based Attacks}
Early adversarial attack research primarily focused on digital world attacks against image classification models\cite{goodfellow2014explaining,madry2017towards,carlini2017towards}. By adding imperceptible perturbations to the original images, classification models could be deceived and make incorrect predictions. The resulting images with added perturbation were known as adversarial examples. 
With the widespread application of adversarial examples, they could be transferred between different architectures, severely affecting the applicability of neural networks. Some efforts have been made to enhance the transferability of adversarial examples by training substitute models, but it was challenging to apply this approach to large models and large datasets. 
Liu et al.\cite{liu2016delving} pioneered the research on the transferability of large-scale models and datasets by introducing an approach based on the ensemble of multiple target models to generate transferable adversarial examples. 
Based on the ensembled model attack, by training an adversarial sample that can deceive multiple attacked models, the success rate of attacks against unknown black-box models will also increase. Zheng et al.\cite{yuan2021meta} introduced the Meta Gradient Adversarial Attack (MGAA) framework, which can be seamlessly ensembled with existing gradient-based attack methods to improve the transferability of adversarial examples across different models. 
Despite the extensive research on adversarial attacks in the field of image classification, there has been a lack of attention on ensemble-based attacks for dense prediction models such as object detection\cite{liang2022large,huang2021rpattack}. 
Meanwhile, most of the existing ensemble model works\cite{huang2019black,lord2022attacking,tashiro2020diversity} simply fuse the outputs of all models, neglecting the differences in model structures, which leads to adversarial optimization mostly heading in a single direction, making it difficult for adversarial examples to converge into a form that is effective for all target models. Recently, Cai et al.\cite{cai2023ensemble} proposed an attack algorithm based on transfer and query combination, demonstrating that adjusting the weights of the ensemble according to the victim model can further improve the attack performance. The universal adversarial perturbations they obtained can successfully attack various object detection models, such as YOLOv3 and Faster R-CNN, but can only be applied to digital images. Our research focused on the application of ensemble models in adversarial patches, 
adjusting weight coefficients based on energy loss feedback from different victim models to generate effective patches for all models.

\subsection{Adversarial Patch Attacks}
The adversarial patch was a special form of adversarial samples and had become a popular research object in the field of adversarial attacks. The adversarial patch covered the local area of the image like a sticker pattern, and the victim models could be successfully deceived simply by placing the patch on the object to be tested\cite{brown2017adversarial}. How to generate adversarial patches with high attack and migration capabilities was the most important problem at this stage. Wu et al.\cite{wu2020making} analyzed the attack performance and transferability of adversarial patches across various datasets, target models, and attack types. Hao et al.\cite{huang2021rpattack} utilized critical pixel locations to add adversarial patches, balance the gradients of both detectors during training, and successfully conducted simultaneous attacks on YOLOv4 and Faster R-CNN object detectors. As object detection continues to rapidly develop, there is a growing body of research that focuses on detecting individuals in the physical world. 
Earlier studies produced adversarial perturbations in digital environments, which were then used to perform physical attacks on actual objects through the creation of adversarial disguises or by directly printing perturbation. 
Thys et al.\cite{thys2019fooling} had devised adversarial patches that are specifically tailored for object detection systems. These patches, when printed on a cardboard surface, resulted in the first-ever implementation of ``stealth'' capabilities for humans under physical world object detection models. Hu et al.\cite{hu2022adversarial} had developed a novel approach to generate scalable adversarial textures that are suitable for attacking object detection models from multiple angles. Their methods involved utilizing a generator and toroidal cropping strategy to create textures that can be applied to surfaces of various shapes. 
In contrast to the aforementioned research, our study is centered on the development of universal adversarial patches and the exploration of their transferability and interpretability.

\section{Methodology}
\label{sec:3}
To generate adversarial patches that exhibit effective attacks against mainstream object detectors and demonstrate good transferability, while retaining attack effectiveness in the physical world through printing on fabrics. We investigate the mechanism of generating adversarial patches from the perspective of an energy function. 
The adversarial patches are first processed by the transformation function and then overlaid on clean samples. The obtained adversarial samples are used as the input of the dynamically optimized ensemble model, and the energy loss output by the ensemble model is ${L_{patch}}$. 
To enhance the effectiveness of adversarial patches, we introduce two additional components: pixel value constraints and smoothness adjustment. The output of Non-Printability Score\cite{sharif2016accessorize} loss ${L_{NPS}}$ and smoothness loss ${L_{smooth}}$ are used as another part of the energy function. 
 By integrating the aforementioned three losses, we derive the overall energy loss and subsequently optimize the adversarial patches via gradient descent to minimize this loss. This approach enables the transferability of the adversarial patch across various object detection models. The overall framework is illustrated in the following Fig.\ref{fig:2}. 
\begin{figure*}[!t]
\centering
\includegraphics[scale=0.25]{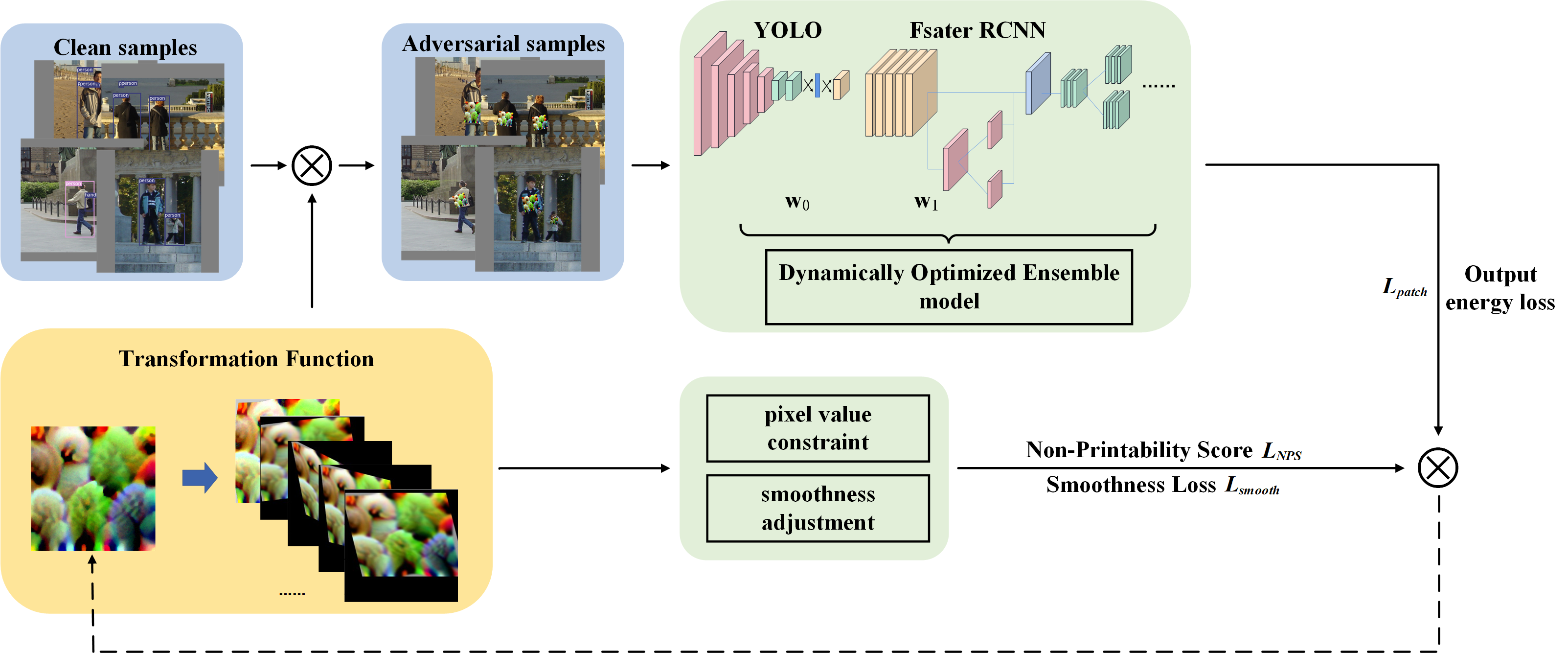}
\caption{Overview of DOEPatch generation framework. The solid line indicates the direction of training the adversarial patch, and the dashed line indicates the direction of backpropagation.}
\label{fig:2}
\end{figure*}

\subsection{Problem Definitions}
Given clean samples ${X}$ with labels ${Y}$, the resulting adversarial examples(${{X_{adv}}{\rm{ = }}X + \delta}$), obtained by adding adversarial perturbation ${\delta}$ onto the images, have the potential to cause the object detection model ${f( \cdot )}$ to make an erroneous prediction. To limit the adversarial perturbation in a reasonable bound, we constrain it within ${\varepsilon}$ by ${{l_\mathbb{P}}}$-norm, the function is expressed as follows, 
\begin{equation}
	f({X_{adv}}) \ne Y,s.t.{\left\| \delta  \right\|_\mathbb{P}} < \varepsilon
	\label{eq:1}
\end{equation}

In contrast to traditional methods of generating adversarial samples, adversarial patches are like stickers that are placed over specific local regions of an image.
In order to investigate the generation mechanism of adversarial patches, we introduce the concept of the energy model and redefine the process of generating adversarial patches for object detection models from an energy perspective. In neural networks, "energy" is an abstract concept that does not directly correspond to energy in physics. In an energy model, "energy" refers to the value of the energy function, which is a description of the system state. When designing adversarial examples, an energy function can be used to quantify the model's response or output distribution. By finding input perturbations that reduce the energy of model outputs (or increase them towards non-target classes), effective adversarial examples can be generated. Here, "energy" can be understood as a measure of the model's confidence or certainty in specific outputs. By optimizing the energy function, attackers can devise inputs that lead to high uncertainty or errors in the model.

The energy function is usually written as $E(X,Y)$, which is used to measure the compatibility of ${X}$ and ${Y}$. It can also be understood as whether ${X,Y}$ matches or not, and the smaller the energy, the higher the match. The construction and training of an energy-based model requires the following parts: firstly, the structure of the energy function is determined, which we define as $\psi =E(G,X,Y),G\in\mathcal{G}$. When ${X,Y}$ are real vectors, ${\mathbf{\psi}}$ can be a simple linear combination, or a set of neural network architectures and weights. When ${X,Y}$ are variable images, sequences of symbols, vectors, or more complex structured objects, ${\mathbf{\varepsilon}}$ can represent a more complex class of functions. Thus, given data $S=\{(X^{i},Y^{i}):i=1...N\}$, we construct the energy function $L(E,S)$ to evaluate how good the current energy output is. For simplicity, this is usually denoted as $L(G,S)$. The purpose of the energy function is to find the parameter ${G}$ that minimizes the energy loss, i.e., $G^{^{*}}=\min_{G\in\mathcal{G}}L(G,S)$. 

Given a clean sample ${X}$ with label ${Y}$, we aim to optimize the adversarial patch such that the person in the image evades recognition by the target detector. In contrast to traditional methods for generating adversarial samples, adversarial patches are like stickers placed on specific localized regions of an image. Therefore, we add adversarial patches to the target objects in the input image by means of a transformation function ${T( \cdot )}$. In order to investigate the generation mechanism of adversarial patches, we introduce the concept of energy model and redefine the process of generating adversarial patches for target detection models from an energy perspective. First, adversarial training is different from the training process of target detection models or energy-based models. The size and distribution of pixel values in the adversarial patch can be viewed as the parameters ${G}$ that need to be updated. Our goal is to construct a suitable energy function to find the optimal adversarial patch minimizing the energy, i.e., $p^{*}=\min_{p}L_{\mathrm{energy}}(p,S)$. The energy function for constructing adversarial patch on energy-based model is defined by the following equation:
\begin{equation}
	L_{\text {energy }}(p, S)=\min \mathbb{E}(f(Y, T(p, X)))+R(p)\label{eq:2}
\end{equation}

The training set ${S}$ is comprised of tuples ${\{ ({X^i},{Y^i}); i = 1...,N\}}$, which ${{X^i}}$ represents the input training example and ${{Y^i}}$ denotes its corresponding ground-truth label. ${p \in {\mathbb{R}^{H \times W \times 3}}}$ denotes the adversarial patch, and ${R(p) = {\left\| p \right\|_\mathbb{P}}{\text{ = (}}\sum\limits_{i,j} {{{\left| {{p_{(i,j)}}} \right|}^\mathbb{P}}} {{\text{)}}^{1/\mathbb{P}}}}$ is the regularization term that restricts the size of each pixel ${{p_{(i,j)}}}$ within the adversarial patch by imposing a ${{l_\mathbb{P}}}$-norm constraint. During the adversarial patch training process, we optimize the confidence scores of candidate bounding boxes belonging to the ``person'' category, which are used as the objective for the optimization algorithm (i.e., the output of ${f(Y,T(p,X))}$). The energy function incorporates the confidence scores of all candidate boxes as part of the optimization process. The adversarial patch serves as the parameter to be updated iteratively until the energy of the target category is minimized through the optimization process. 

\subsection{Generating Adversarial Patch}
{\bf{Transformation Function:}}
To facilitate the generalization of the adversarial patch from the digital to the physical world, we need to consider the interference of physical world factors on the patches during the training process.  Therefore, based on the idea of Expectation over Transformation(EoT)\cite{athalye2018synthesizing} framework, we design the function ${{T}( \cdot )}$ to achieve random rotation and translation of the adversarial patch within a certain range to simulate the transformation of the shooting perspective in the physical world. At the same time, we propose the function ${{T}_{l}( \cdot )}$ to simulate the lighting transformation in the real world by adding random noise to the whole image and adjusting the brightness and contrast.In real-world physical environments, the scale of the patch tends to decrease as distance increases and the resulting imagery becomes more blurred. We introduce a novel distance transformation function. We employ a Gaussian kernel to blur the image, ensuring that the smaller the target object becomes, the larger the convolution kernel becomes, resulting in a more blurred adversarial patch image. Our goal is to develop an adversarial patch pattern that can be printed on a T-shirt to enable a person to evade object detection models. To achieve this goal, we introduce the Thin Plate Splines(TPS)\cite{donato2002approximate} approach to simulate non-rigid deformations such as folds on the T-shirt's surface. Using the aforementioned transformation methods, we devise a transformation function ${{T_p}( \cdot )}$ to apply the adversarial patch to the target image, 
\begin{equation}
\begin{gathered}
  {X_{adv}} = {T_p}(p,X) =  \hfill \\
  {T_{l}}(((1 - {M_p}) \cdot X + {T}({T_d}({M_p} \cdot {T_{TPS}}(p)))))
  \label{eq:3} \hfill \\ 
\end{gathered} 
\end{equation}

${X_{adv}}$ is the image after applying the counter patch, ${{M_p} \in {[0,1]^d}}$ represents the mask region of the adversarial patch on the target image, which is determined based on the bounding box obtained from the label. The ${{T_p}( \cdot )}$ function encompasses four main types of transformations, of which the first one involves applying TPS transformation ${T_{TPS}( \cdot )}$ to the generated adversarial patch. This serves to introduce distortions, thereby simulating non-rigid deformations on clothing surfaces. Subsequently, based on the size of the bounding box in the input image, the distance is estimated and a distance transformation function ${{T_d}( \cdot )}$ is applied to the generated adversarial patch. This enables the patch image to be blurred to varying degrees.The perspective transformation function ${{T}( \cdot )}$ is then used to simulate the perspective transformation during shooting in the physical world. In the last step, the illumination transformation ${T_{l}( \cdot )}$ is applied to the whole image to simulate the illumination transformation phenomenon in the physical world to further enhance the robustness of the adversarial patch.
 
\begin{figure}[!t]
\centering
\includegraphics[scale=0.32]{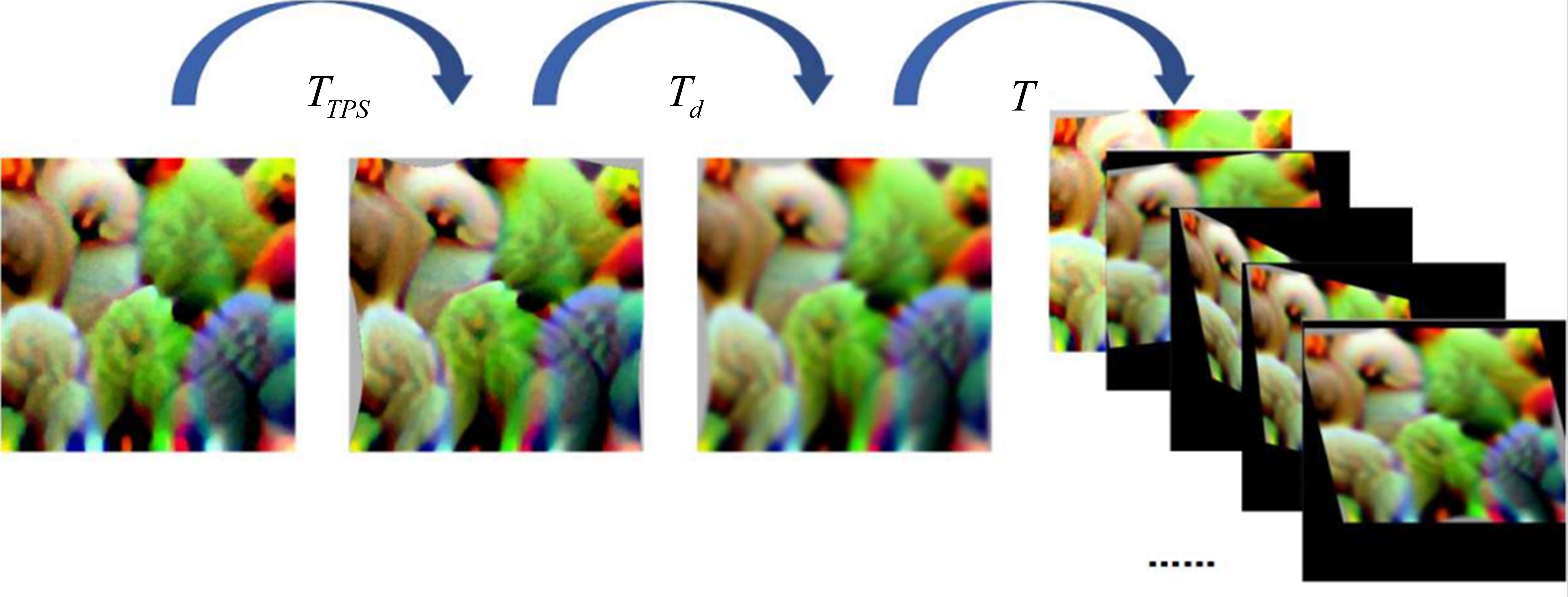}
\caption{The transformation process of the adversarial patch.}
\label{fig:3}
\end{figure}

{\bf{Adversarial Loss:}}
For contemporary object detection models, it is crucial to design an appropriate energy function that can effectively deactivate the detection of the person category by minimizing the energy. Given images ${X}$ and the ground-truth labels ${Y}$, contemporary object detectors take this information as input, and utilize  the detector's prediction function ${{f}( \cdot )}$ to generate bounding boxes' information. The information includes the information of the position of the candidate box, the confidence scores of the target, and the scores of each category. The candidates with confidence scores exceeding a predefined threshold are selected as the predicted results. Consequently, we incorporate the expected confidence scores of the output as a part of the energy function for adversarial patch optimization, aiming to reduce the target's confidence scores by optimizing the adversarial patch. Specifically, we begin by randomly initializing the adversarial patch and utilizing the transformation function ${{T_p}( \cdot )}$ to apply the patch to the original image, which serves as the input for the object detection models. With the function ${{f}( \cdot )}$, we can get the information about the ${N}$ candidate boxes output by the detector, and ${f_j}$ denotes the information of the ${j}$ candidate box from output. Previous research incorporated the maximum confidence score of the target category, which was output by the detection network, as a constituent element in the energy function. Unlike previous work\cite{thys2019fooling,song2018physical}, we consider all ${N}$ bounding boxes output by the detection model as optimization targets during each iteration, taking into account changes of all confidence scores comprehensively. To reduce the computational cost, we select candidate boxes whose confidence scores for the category ``person'' are higher than the set threshold value ${\mu }$ as attack objects. Subsequently, we continually optimize and update the patch pixels through gradient descent. In the case of one-stage object detection models, such as YOLOv2, YOLOv3, the detection network directly outputs all bounding boxes and associated category information. We integrate confidence scores of candidate boxes relevant to the ``person'' category as a constituent of the energy function. For two-stage object detection models like Faster R-CNN, person-related bounding boxes are extracted through Region Proposal Network(RPN), incorporating both positive and negative sample information. Positive candidate boxes with confidence scores surpassing a threshold are selectively chosen as attack targets. Hence, we compute the confidence score of all bounding boxes in the adversarial sample that exceeds the threshold ${\mu }$ of the target detection network. This score is merged into the energy function used for training the adversarial patch. Through iterative optimization, the patch is fine-tuned to minimize the total energy loss, as defined in Eq.\ref{eq:4}. The generative algorithm, designed based on the energy function, optimizes the adversarial patch during training, thereby reducing energy and causing a malfunction in the target detector's person detection capability. By training a highly complex adversarial patch, the energy of the remaining categories is elevated, leading to a reduction in the energy associated with the person category. Sec.\ref{sec:4.4} of the experiment provides a more detailed explanation regarding the impact of the adversarial patch on the energy levels of various categories. 
\begin{equation}
{L_{obj}} = \mathop {\min \mathbb{E}}\limits_p \left(\sum\limits_{j = 1}^N {\max ({f_j}(Y,{T_p}(p,X)),\mu )} \right)\label{eq:4}
\end{equation} 

\subsection{Dynamically Optimized Ensemble Model}
\label{sec:doem}
To achieve person invisibility across multiple object detection models, the generated adversarial patch effectively impacts the predictive information of all detection algorithms, resulting in a significant reduction of the energy associated with the person category in the output. For ${K}$ object detection models, ${f( \cdot )}$ denotes its outputs and 
${\mathbf{w}=[{{\text{w}}_1},{{\text{w}}_2},...,{{\text{w}}_K}]}$ denotes the weights used to adjust different models, subject to constraint ${\sum\limits_{i = 1}^K {{{\text{w}}_i}}  = 1}$. In traditional approaches, the training is usually done by manually adjusting each value, or by directly averaging the values. The ensemble model for the adversarial patch can be expressed as follows, 
\begin{equation}
\min_p\sum_{i=1}^K\mathrm{w}_i\cdot\left(\mathbb{E}(\sum_{j=1}^N\max(f_j^i(Y,T_p(p,X)),\mu)\right) \label{eq:5}
\end{equation} 

${f_{j}^{i}(\cdot)}$ denotes the information of the ${j\mathrm{-th}}$ candidate box output in the ${i\mathrm{-th}}$ detection model.The attack is achieved by optimizing the adversarial patches so that the outputs of all models are reduced. 
However, it is challenging to achieve an ideal balance point for the weights in object detection models with vastly different structures such as YOLO and Faster R-CNN, and overfitting is therefore prone to occur during the training process. Sec.\ref{sec:4.5} of this paper provides a detailed analysis of this phenomenon. To address the aforementioned challenges, we propose a dynamically optimized ensemble model for multiple object detection models based on the Min-Max adversarial training approach. During the training process, the proportions of each detection model's contribution to the overall energy output are dynamically adjusted through parameter tuning to achieve the universality of the adversarial patch, as calculated in Eq.\ref{eq:6}. The training procedure involves an outer minimization phase, with fixed parameters ${\mathbf{w}}$, where the adversarial patch is optimized through backpropagation to minimize the energy output of the ensemble model. And an inner maximization phase is performed with a fixed adversarial patch, using projected gradient descent to balance the weights ${\mathbf{w}}$ of different models and achieve the maximum energy output of the ensemble model. 
However, excessive parameter fluctuations can lead to oscillations during the training process, making it difficult to converge to an optimal adversarial patch. On the other hand, if the parameter fluctuations are too small, the loss may stop decreasing after reaching a certain point, failing to reach the desired level. 
In order to stabilize the process of the weights ${\mathbf{w}}$, the dynamic rate adjustment parameter $\nu$ is increased to control the speed of ${\mathbf{w}}$ change. The value of the parameter $\nu$ was verified in Sec.\ref{sec:4.1}.

This method of using a single encoding can compromise the generalization capabilities across other models, which can lead to instability during the training process due to potential training bias towards a specific model \cite{lu2019block}. Thus, to alleviate this issue, it is common to introduce a strong concave regularizer \cite{qian2019robust} in the inner maximization step. 
\begin{equation}
{L_{patch}} = \mathop {\min }\limits_{p} \{ \mathop {\max }\limits_{{\mathbf{w}} \in W} \sum\limits_{i = 1}^K {{{\text{w}}_i}L_{obj}^i}  - L({\mathbf{w}})\}\label{eq:6}
\end{equation}

The weight parameters ${\mathbf{w}}$ are defined as a probability simplex, satisfying ${W{\text{ = \{ w\textbar}}{{\mathbf{1}}^T}{\mathbf{w = 1,}}{{\text{w}}_i} \in {\text{[0,1]\} }}}$. $L_{obj}^i = \mathbb{E}\left(\sum\limits_{j = 1}^N \max (f_j^i(Y,{T_p}(p,X)),\mu) \right)$ denotes the output of the $j$-th object detector, ${L({\mathbf{w}}) = \frac{\gamma }{2}\left\| {{\mathbf{w}} - \frac{1}{K}} \right\|_2^2}$ represents the regularization term, and ${\gamma  > 0}$ is the regularization parameter. If ${\gamma  = 0}$, the adversarial patch is designed by training under the maximum attack loss(worst-case attack scenario). If ${\gamma  \to \infty}$, the inner maximization corresponds to computing the average of ${K}$ attack losses. As such, the regularization parameter serves to strike a balance between the maximum loss attacking strategy and the average attacking strategy, thereby enhancing the attack efficacy of the adversarial patch across all targeted detection models.

The training process is illustrated as Algorithm~\ref{ALG:1}. 
In line 1 we initialize the adversarial patch, parameters, and model.  Then, load the dataset and the corresponding labels in line 3. In line 4, we overlay the adversarial patch $p$ onto the original sample $x$ through the transformation function ${{T_p}( \cdot )}$ to obtain the adversarial sample ${x_{adv}}$. The 5-6 lines enter the external minimization stage, the fixed weight parameter ${\mathbf{w}}$ is the value in the ($t$-1) state, and the total output energy loss of different target detection models is calculated. In line 7, we update the adversarial patch through backpropagation to minimize the total output energy loss. The 8 line is the internal maximization stage, the fixed parameter $p$ is the value in the ($t$) state, and ${\mathbf{w}}$ is updated by projected gradient descent to maximize the total output energy loss. Finally, repeat the Min-Max confrontation training until the maximum number of iterations is reached.

\begin{algorithm}[!ht]
    \caption{ Dynamically optimized ensemble model}
    \label{alg:DOE}
    \renewcommand{\algorithmicrequire}{\textbf{Input:}}
    \renewcommand{\algorithmicensure}{\textbf{Output:}}
    
    \begin{algorithmic}[1]
        \REQUIRE clean samples $X$, ground-truth labels $Y$, Physical transformation $T$, $K$ object detection models $F = \{ {f^1},...,{f^K}\}$, bounding boxes output by the detection model $N$.
        \ENSURE  adversarial patch ${p_{adv}}$   
        
        \STATE  Initializing object detection models $F$, setting the initial weight parameters ${{\mathbf{w}}^{(0)}}$, regularize coefficient $\gamma$, dynamic rate adjustment parameter $\nu$, initializing adversarial patch $p$ with random values.

        \FOR{the max epochs}
            \STATE The mini-batch of samples consisting of $n$ samples, $x \in X$, $y \in Y$
            \STATE adversarial samples ${x_{adv}} = {T_p}(p,x)$
            \STATE outer minimization: fixing ${\mathbf{w}} = {{\mathbf{w}}^{(t - 1)}}$,
            \STATE calculating $\sum\limits_{i = 1}^K {{{\text{w}}_i}\{ } \mathbb{E}(\sum\limits_{j = 1}^N {\max (f_j^i(Y,{T_p}(p,X)),\mu )} \}$  
            \STATE updating $p$ by minimizing through backpropagation
            \STATE inner maximization: fixing $p = {p^{(t)}}$, updating weight ${{\mathbf{w}}^{(t)}}$ through projected gradient descent, ${{\mathbf{w}}^{(t)}} = pro{j_W}({{\mathbf{w}}^{(t - 1)}} + \nu {\nabla _{\mathbf{w}}}(\sum\limits_{i = 1}^K {{{\text{w}}_i}L_{obj}^i}  - R({\mathbf{w}})))$
        \ENDFOR
    \end{algorithmic}
    \label{ALG:1}
\end{algorithm}

\subsection{Overall Optimization Process}
We aim to generate universal adversarial patches that can be effective in both the digital and physical world, with a focus on optimizing the patch's pixel values and smoothness. The adversarial patch pixel value constraint component optimizes the color disparity between adversarial patches and regular printers by evaluating the Non-Printable Score. This definition is reflected mathematically in the following equation,

\begin{equation}
{L_{NPS}} = \sum\limits_{{i_{patch}} \in p} {\mathop {\min }\limits_{{c_{print }} \in C} \left| {{i_{patch}} - {c_{print}}} \right|} \label{eq:7}
\end{equation}

${c_{print}}$ represents a printable color and ${i_{patch}}$ represents a pixel in the adversarial patch $p$. Devices such as printers and screens can only reproduce a subset of the ${[0,1]^3}$ RGB color space.
Furthermore, to ensure the smoothness of generated adversarial patches, the smoothness adjustment component calculates the minimization of smoothness loss (${L_{smooth}}$) for the adversarial patches.

\begin{equation}
{L_{smooth}} = \sum\limits_{i,j} {{{({{({p_{i,j}} - {p_{i + 1,j}})}^2} + {{({p_{i,j}} - {p_{i,j + 1}})}^2})}^{\frac{1}{2}}}}\label{eq:8}
\end{equation}

${p_{i,j}}$ refers to the pixel value at the coordinate $(i,j)$ in the adversarial patch $p$. When adjacent pixel values are close, the smoothness loss ${L_{smooth}}$ is low (i.e., perturbation is smooth). Therefore, by minimizing ${L_{smooth}}$, the smoothness of the perturbed image can be increased, enhancing the feasibility of physical attacks. Finally, in order to generate adversarial patches that are effective in both the physical and digital worlds, we define the total training energy loss ${L_{attack}}$ as follows. 
\begin{equation}
{L_{attack}} = {L_{patch}} + \alpha  \cdot {L_{NPS}} + \beta  \cdot {L_{TV}}\label{eq:9}
\end{equation}

$\alpha $ and $\beta$ denote weight coefficients determined empirically. The goal of the optimization process is to minimize the overall energy function ${L_{attack}}$ by jointly minimizing its constituent terms. 
\section{Experiment}
\label{sec:4}
\subsection{Implementation Details}
\label{sec:4.1}
To conduct experiments on digital image attacks, we utilized the Inria Person dataset\cite{dalal2005histograms} consisting of 614 person images for training and 288 images for testing purposes. This dataset had been used extensively in research on adversarial attacks against the person category, such as the work of Thys et al.\cite{thys2019fooling} and Hu et al.\cite{hu2022adversarial}. To evaluate the attack effectiveness and transferability of adversarial patches, we first measure their impact by computing the Average Precision (AP) values of the images with and without the patches. The larger the drop in AP, the stronger the attack effectiveness. In the physical experiments, to better evaluate the attack effectiveness of adversarial patches, we introduced the evaluation metric Attack Success Rate (ASR)\cite{xu2020adversarial}. ASR is defined as the proportion of images successfully attacked among all test images. Considering the existing baseline work, most of them are based on YOLOv2, Faster R-CNN for verification. Therefore, for a fair comparison, we designed our adversarial patches to attack mainly the mainstream single-stage object detection models, such as YOLOv2, YOLOv3, and YOLOv4. For more complex two-stage object detection models, we chose the Faster R-CNN detection model based on the backbone of the VGG16 and ResNet50 architectures. All detection models were trained on the COCO dataset, with the output confidence threshold set to 0.5 and the Non-Maximum Suppression (NMS) threshold to 0.4 during the inference stage. Due to laboratory arithmetic limitations, we only validated 2-3 combinations of target detectors in the ensemble model. Therefor, we consider three different sets of object detection models to design ensemble models for the adversarial patches: 
\begin{itemize}
    \item DOEPatch(YY):  Adversarial patch for ensemble model training composed of YOLOv2 and YOLOv3.
    \item DOEPatch(YF):  Adversarial patch for ensemble model training composed of YOLOv3 and Faster-VGG16.
    \item DOEPatch(YYF): Adversarial patch for ensemble model training composed of YOLOv2, YOLOv3 and Faster-VGG16.
\end{itemize}

And based on the designed Energy model (EmPatch), we retrained adversarial patches: 
\begin{itemize}
    \item EmPatch(Y1): Adversarial patch for single model YOLOv2.
    \item EmPatch(Y2): Adversarial patch for single model YOLOv3.
    \item EmPatch(F): Adversarial patch for single model Faster-VGG16.
\end{itemize}

\begin{figure}[!t]
\centering
\includegraphics[scale=0.28]{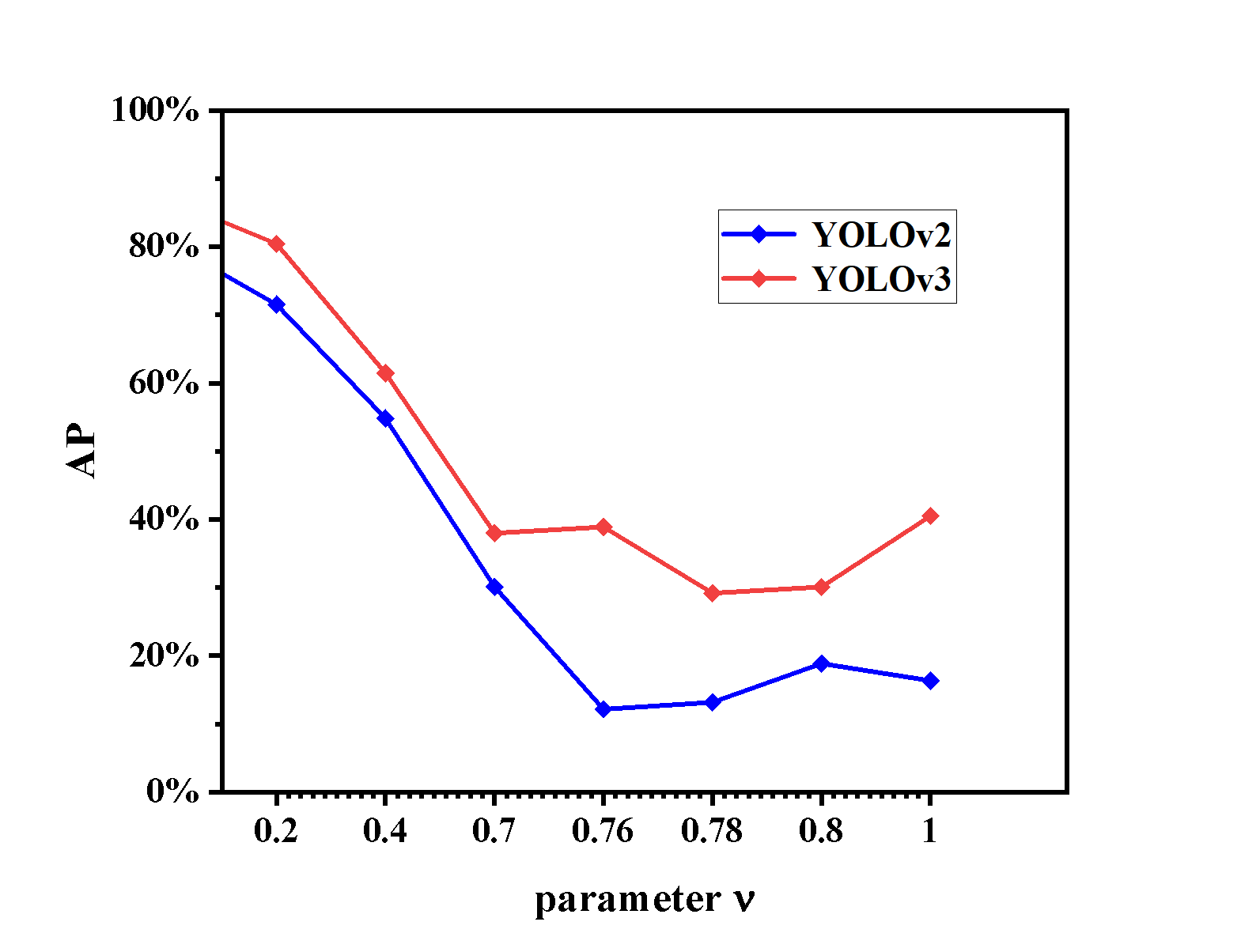}
\caption{The process of AP value changes with the variation of the internal dynamic rate adjustment parameter $\nu$.}
\label{fig:4}
\end{figure}

 The entire algorithm is designed in PyTorch and trained on an NVIDIA RTX 3090 24GB GPU cluster. The adversarial patches are set to a size of $3\times300\times300$, optimized using the Adam\cite{kingma2014adam} optimizer, and the learning rate for external minimization optimization is set to 0.03. 
 We determined that the range of $\alpha$ parameter is in [0,0.05] and the range of $\beta$ parameter is in [0,0.5] based on the experience accumulated in previous related work. Then, by analyzing the generated adversarial images, we found that the gap between the patch when it is brighter and the printed effect is bigger, and the $\alpha$ parameter needs to be tuned down. If the noise on the patch is larger, it indicates that the image is not smooth enough and the  parameter $\beta$ needs to be increased. Based on the above phenomenon, we determine that the parameters $\alpha  = 0.01$, $\beta=0.165$, $\mu=0.4$ and the regularization coefficient was set $\gamma=0.1$. The internal dynamic rate adjustment parameter $\nu$ was determined based on experimental results on DOEPatch(YF), as shown in Fig.\ref{fig:4}. Validation showed that with $\nu=0.78$, the adversarial patch achieved good attack effects on both YOLOv2 and YOLOv3.
\subsection{Digital Attack Evaluations}
\label{sec:4.2}
\begin{figure*}[!t]
\centering
\includegraphics[scale=0.32]{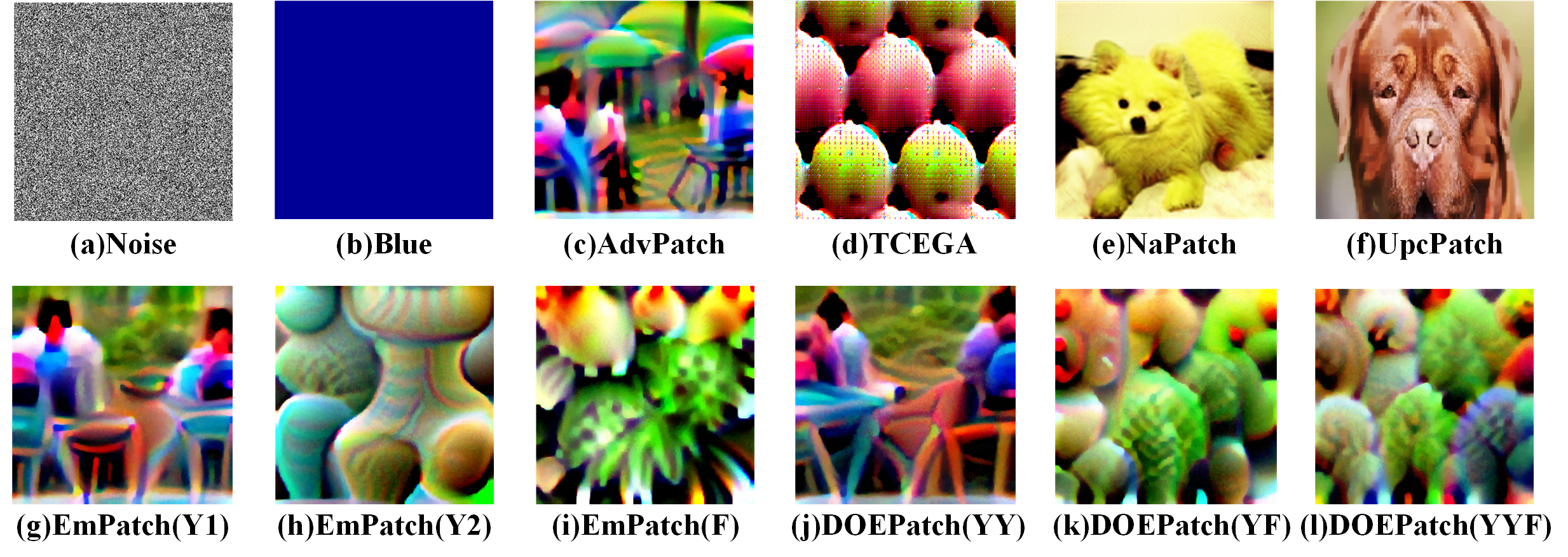}
\caption{(a)-(f) show the comparative adversarial patches, while (g)-(i) show the adversarial patches generated by us for YOLOv2, YOLOv3, and Faster R-CNN, respectively. (j)-(l) show the adversarial patches generated based on three sets of ensemble models.}
\label{fig:5}
\end{figure*}

To more fully demonstrate the superiority of the adversarial patch based on dynamically optimized ensemble model design in terms of attack performance and transferability, this study added six comparative experiments to compare with related work that has publicly released adversarial patch results in recent years. Fig.\ref{fig:5}(a) is an adversarial patch constructed using random noise, Fig.\ref{fig:5}(b) is a pure color adversarial patch designed as Blue, Fig.\ref{fig:5}(c) is an adversarial patch called AdvPatch designed by Thys et al.\cite{thys2019fooling} for the YOLOv2 algorithm, Fig.\ref{fig:5}(d) is an adversarial patch named TCEGA designed by Hu et al.\cite{hu2022adversarial} for the YOLOv2 algorithm, Fig.\ref{fig:5}(e) is an adversarial patch designed based on StyleGAN to attack the YOLOv2 algorithm named NaPatch\cite{hu2021naturalistic}, and Fig.\ref{fig:5}(f) is an adversarial patch designed for the Faster R-CNN algorithm named UpcPatch\cite{huang2020universal}. Based on the energy function reconstruction algorithm for adversarial patch generation, this study designed adversarial patches EmPatch(Y1), EmPatch(Y2), and EmPatch(F) respectively for YOLOv2, YOLOv3, and Faster R-CNN, as shown in Fig.\ref{fig:5}(g), Fig.\ref{fig:5}(h), and Fig.\ref{fig:5}(i), respectively. Adversarial patches designed based on the dynamically optimized ensemble model, DOEPatch(YY), DOEPatch(YF), and DOEPatch(YYF), are shown in Fig.\ref{fig:5}(j), Fig.5k, and Fig.\ref{fig:5}(l), respectively. Note that AdvPatch and TCEGA were trained on the same dataset and used the same YOLOv2 model and parameters as those in this study, making them highly comparable. However, NaPatch and UpcPatch focused more on the stealthiness of the adversarial patches and exhibit certain differences compared to this study in terms of the dataset used. 

All adversarial patches were applied to the images in the Inria test set, and the effectiveness of the attacks was evaluated by calculating the changes in the AP of the person category in the output of different object detection models. Experimental results in the digital world are shown in Tab.\ref{lab:1}, where ``Clean'' represents the AP value of the person category in the original test set. The AP values for the test sets with 12 additional adversarial patches were tested using 5 mainstream object detection models. Under the detection of YOLOv2, DOEPatch(YY) reduced the AP value by 67.49\%, slightly lower than AdvPatch. However, the attack performance of DOEPatch(YY) is better than other adversarial patches such as TCEGA. Meanwhile, the best attack performance was achieved under the detection of YOLOv3, where DOEPatch(YY) was able to reduce the AP value to 29.20\%. Our energy-based adversarial patch EmPatch(Y2) achieved an AP reduction of 54.11\% under the detection of YOLOv3, and its attack performance was far superior to the test results of other patches. For the more complex Faster-VGG detection algorithm, DOEPatch(YF) and DOEPatch(YYF) achieved the highest attack effects, reducing AP values by 43.94\% and 44.81\%, respectively. They also demonstrated strong attack effectiveness under the detection of YOLOv3, reducing AP values by 41.91\% and 41.65\%, respectively. Due to differences in the dataset, UpcPatch only reduced the AP value by 6.10\% in the case of Faster R-CNN. 

To comprehensively compare the transferability of adversarial patches, we conducted black-box attack tests on YOLOv4 and Faster-Res detection algorithms. Our trained DOEPatch(YY) can reduce the AP value by 50.55\% in the detection of YOLOv4, while AdvPatch and TCEGA can only reduce it by 29.14\% and 28.21\%, respectively. Under Faster-Res, the attack performance was generally poor, but DOEPatch(YY) and DOEPatch(YF) still achieved the best attack results, reducing the AP value by 18.55\% and 17.05\%, respectively. The ensemble model we designed incorporates commonly used networks in object detection, such as ResNet, VGG, and others. During the training process, we ensured that adversarial patches could disrupt common feature extraction networks, thus affecting the recognition of object detectors, achieving a certain degree of generality. Moreover, the networks used in common target detectors will not be completely different. Therefore, it will also be effective in attacking against models other than the available ones. Overall, the adversarial patches generated based on a dynamically optimized ensemble model exhibited strong attack performance across multiple targeted models and demonstrated favorable transferability by causing significant output interference against unknown black-box models.

\begin{table*}[!ht]
    \caption{The experiment in the digital world measured the attack performance of various adversarial patches against targeted detectors, with changes in AP values used as the metric. The left axis lists all the adversarial patches, and the top axis lists the object detectors targeted for attack, the best case is in bold.}
    \centering
    \begin{tabular}{l|l|l|l|l|l}
    \hline
        ~ & YOLOv2(drop) & YOLOv3(drop) & Faster-VGG(drop) & YOLOv4(drop) & Faster-Res(drop) \\ \hline
        Clean & 80.68\%(-) & 87.00\%(-) & 91.04\%(-) & 94.60\%(-) & 90.53\%(-) \\ \hline
        Noise & 76.85\%(3.83) & 85.95\%(1.05) & 87.32\%(3.72) & 91.67\%(2.93) & 83.37\%(7.16) \\ \hline
        Blue & 76.94\%(3.74) & 85.28\%(1.72) & 87.68\%(3.36) & 90.85\%(3.75) & 85.89\%(4.64) \\ \hline
        AdvPatch\cite{thys2019fooling} & \textbf{12.59\%(68.09)} & 70.08\%(16.92) & 65.23\%(25.81) & 65.46\%(29.14) & 74.17\%(16.36) \\ \hline
        TCEGA\cite{hu2022adversarial} & 15.46\%(65.22) & 57.43\%(29.57) & 68.23\%(22.81) & 66.39\%(28.21) & 76.20\%(14.33) \\ \hline
        NaPatch\cite{hu2021naturalistic} & 54.71\%(25.97) & 74.88\%(12.12) & 83.35\%(7.69) & 87.54\%(7.06) & 83.11\%(7.42) \\ \hline
        EmPatch(Y1) & 15.02\%(65.66) & 59.73\%(27.27) & 61.06\%(29.98) & \textbf{49.49\%(45.11)} & 73.83\%(16.70) \\ \hline
        EmPatch(Y2) & 65.99\%(14.69) & \textbf{32.89\%(54.11)} & 85.70\%(5.34) & 84.97\%(9.63) & 85.82\%(4.71) \\ \hline
        DOEPatch(YY) & \textbf{13.19\%(67.49)} & \textbf{29.20\%(57.80)} & 60.91\%(30.13) & \textbf{44.05\%(50.55)} & \textbf{71.98\%(18.55)} \\ \hline
        EmPatch(F) & 75.56\%(5.12) & 80.28\%(6.72) & 50.97\%(40.07) & 86.61\%(7.45) & 78.16\%(12.37) \\ \hline
        UpcPatch\cite{huang2020universal} & 76.42\%(4.26) & 84.06\%(2.94) & 84.94\%(6.10) & 90.96\%(3.64) & 85.28\%(5.25) \\ \hline
        DOEPatch(YF) & 59.64\%(21.04) & 45.09\%(41.91) & \textbf{47.10\%(43.94)} & 74.23\%(20.37) & \textbf{73.48\%(17.05)} \\ \hline
        DOEPatch(YYF) & 27.90\%(52.78) & 45.35\%(41.65) & \textbf{46.23\%(44.81)} & 70.85\%(23.75) & 74.95\%(15.58) \\ \hline
    \end{tabular}
    \label{lab:1}
\end{table*}

\subsection{Physical Attack Evaluations}
\label{sec:4.3}
\begin{figure}[!t]
	\centering
	\includegraphics[scale=0.16]{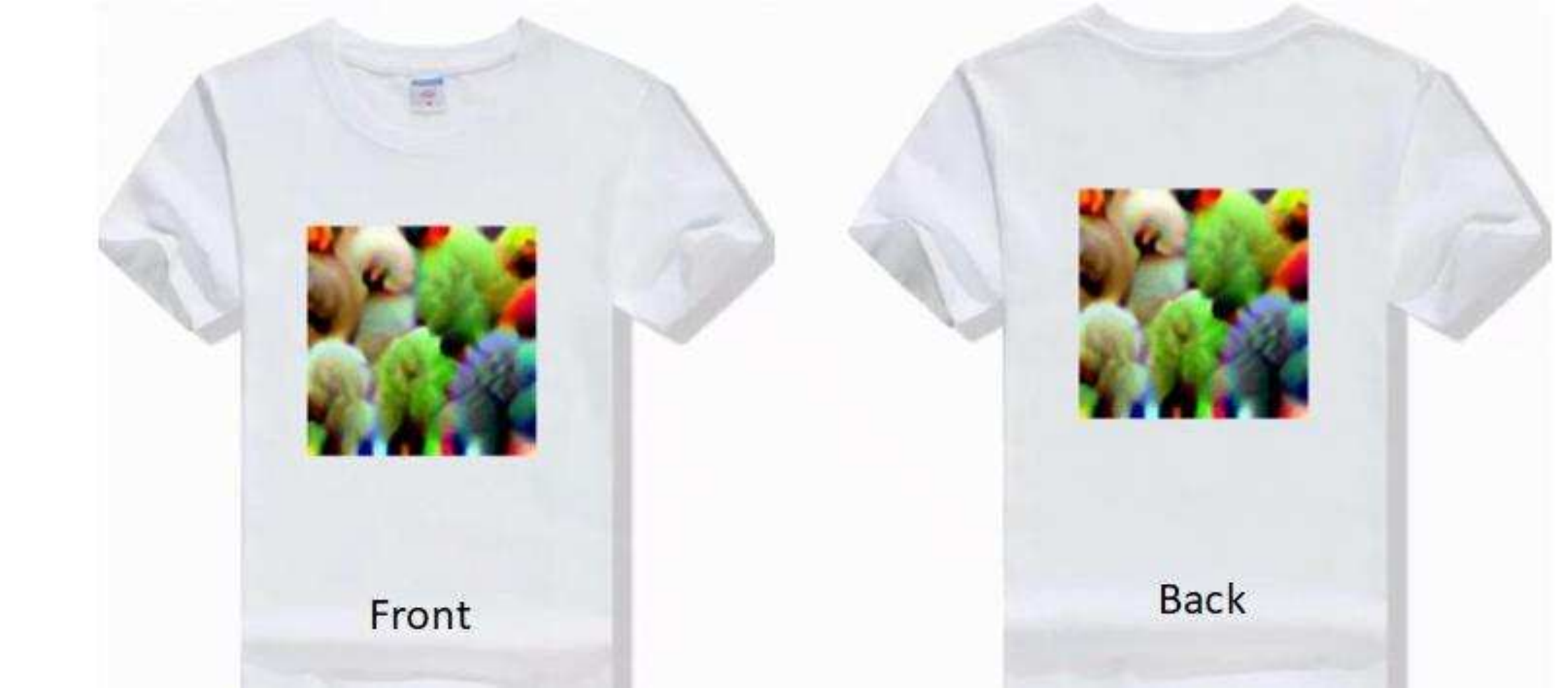}
	\caption{Real-world adversarial T-shirts.}
	\label{fig:6}
\end{figure}

To validate the attack efficacy of the adversarial patches in the physical world, we printed them on T-shirts to simulate the target detection process in real-world settings. Specifically, the adversarial patch in the digital world consists of 300$\times$300 pixels, which we enlarged to 30cm$\times$30cm and printed the adversarial patch on a white cotton T-shirt via DTF-400A printer for testing in the physical world, as shown in Fig.\ref{fig:6}. Due to time and financial constraints, we have only produced three sets of counter patches, DOEPatch(YY), DOEPatch(YF), and DOEPatch(YYF). In order to reduce the pixel difference between the digital and physical worlds, based on the work of Mahmood Sharif et al.\cite{sharif2016accessorize}, we start by making a colour palette containing a one-fifth of the RGB colour. Then, after making it through this printer, the image is captured with the camera of the Redmi K50 Pro to obtain a set of RGB triplets printed in the physical world. In order to reduce the computational complexity, we convert this set of colour sets and into a set of 30 RGB triples that have the smallest change in distance from the original set of colours, i.e., we obtain the printable RGB triple C in Eq.\ref{eq:7}.

\begin{figure}[!t]
	\centering
	\includegraphics[scale=0.26]{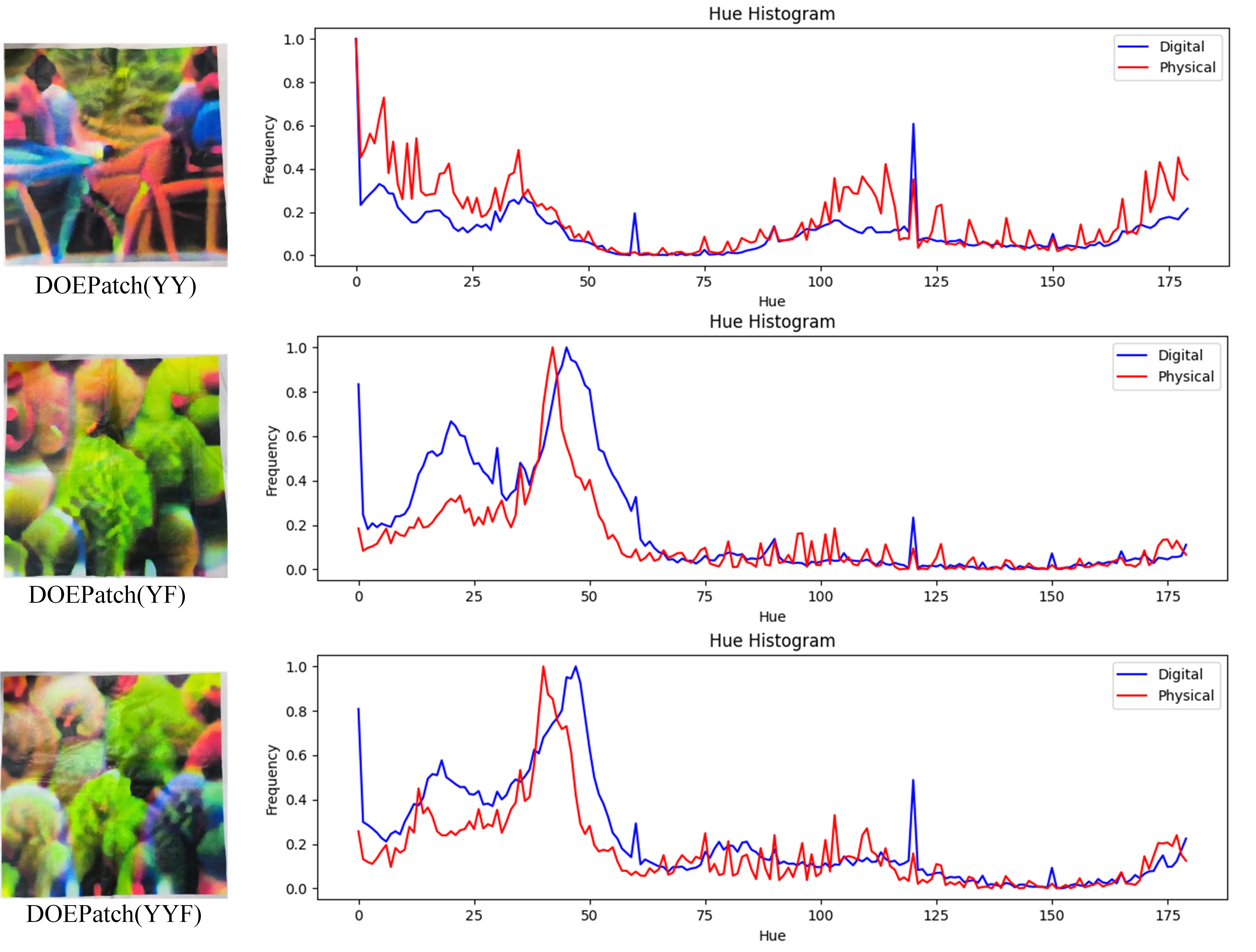}
	\caption{The distribution of adversarial patch tones in both digital and physical worlds.}
	\label{fig:nps}
\end{figure}

In order to evaluate the difference in pixel values of the adversarial patch after printing, we evaluate it by calculating the histogram distribution of the hue channels in the HSV colour space. As shown in Fig.\ref{fig:nps}, the image on the left is the adversarial patch after printing, and the curves on the right represent the variation of the patch's hue distribution in the digital and physical worlds. We can find that the change curves in the digital and physical worlds basically overlap, and the similarity of the hue distributions in the adversarial patches DOEPatch(YY), DOEPatch(YF), and DOEPatch(YYF) are computed to be 0.8235, 0.8332, and 0.8207, respectively. In computing the histogram similarity, the range of the results is usually in [-1 ,1], with 1 indicating a perfect match and -1 indicating a complete mismatch. Therefore, we have greatly reduced the colour loss of digital to physical world transfer by the Non-Printable Score optimized adversarial patches.

Regarding the smoothness of the adversarial patches, we directly calculate the loss of smoothness of the adversarial patches in the digital and physical worlds by using the Eq.\ref{eq:8}. The experimental results are shown in Tab.\ref{lab:smooth}. The overall smoothness of the adversarial patch in the physical world is lower than that in the digital world. The printer, due to the limitation of RGB colour space, some of the colours in the digital space cannot be printed out accurately, which makes the smoothness loss lower. The biggest decrease in the DOEPatch(YYF) reaches 0.00077, which also shows that the pixel values of the counterpatch after printing have very little change and have better smoothness. 
\begin{table*}[!ht]
	\caption{Smoothness values of adversarial patches in the digital and physical worlds.}
	\centering
	\begin{tabular}{c|c|c|c|c|c|c}
		\hline
		~ & \multicolumn{3}{c|}{Digital} & \multicolumn{3}{c}{Physical} \\ \hline
		Methods & DOEPatch(YF) & DOEPatch(YY) & DOEPatch(YYF) & DOEPatch(YF) & DOEPatch(YY) & DOEPatch(YYF) \\ \hline
		$L_{smooth}$ & 0.01333 & 0.01327 & 0.01332 & 0.01278 & 0.01307 & 0.01255 \\ \hline
	\end{tabular}
	\label{lab:smooth}
\end{table*}

Due to time and venue constraints, we conducted physical experiments in four different scenarios, including indoor and outdoor environments, under various lighting conditions. In each scenario, two pedestrians of different body sizes were arranged. During each data collection session, one of the individuals wore the adversarial T-shirt. Firstly, to verify the transferability of adversarial patches, we collected four sets of videos for each adversarial patch within a distance range of 2 to 4 meters and an angle range of -45° to 45°. By extracting images from the videos, each adversarial patch obtained 160 test data samples. The experimental results, as shown in Tab.\ref{lab:physical}, demonstrate that DOEPatch (YY) achieved the highest attack effectiveness in YOLOv2, reaching 56.87\%. Our adversarial patches also demonstrated some attack effectiveness when confronted with the more complex Faster-VGG detection model. 
\begin{table*}[!ht]
	\caption{ASR test results of adversarial patches under different detectors.}
	\centering
	\begin{tabular}{c|c|c|c|c|c}
		\hline
		~ & YOLOv2 & YOLOv3 & Faster-VGG & YOLOv4 & Faster-Res \\ \hline
		DOEPatch(YY) & 56.87 & 33.75 & 21.25 & 30.62 & 15.62 \\ \hline
		DOEPatch(YF) & 35.00 & 28.12 & 13.75 & 25.00 & 8.75 \\ \hline
		DOEPatch(YYF) & 43.75 & 20.62 & 16.87 & 21.87 & 6.87 \\ \hline
	\end{tabular}
	\label{lab:physical}
\end{table*}

\begin{figure*}[!t]
\centering
\includegraphics[scale=0.46]{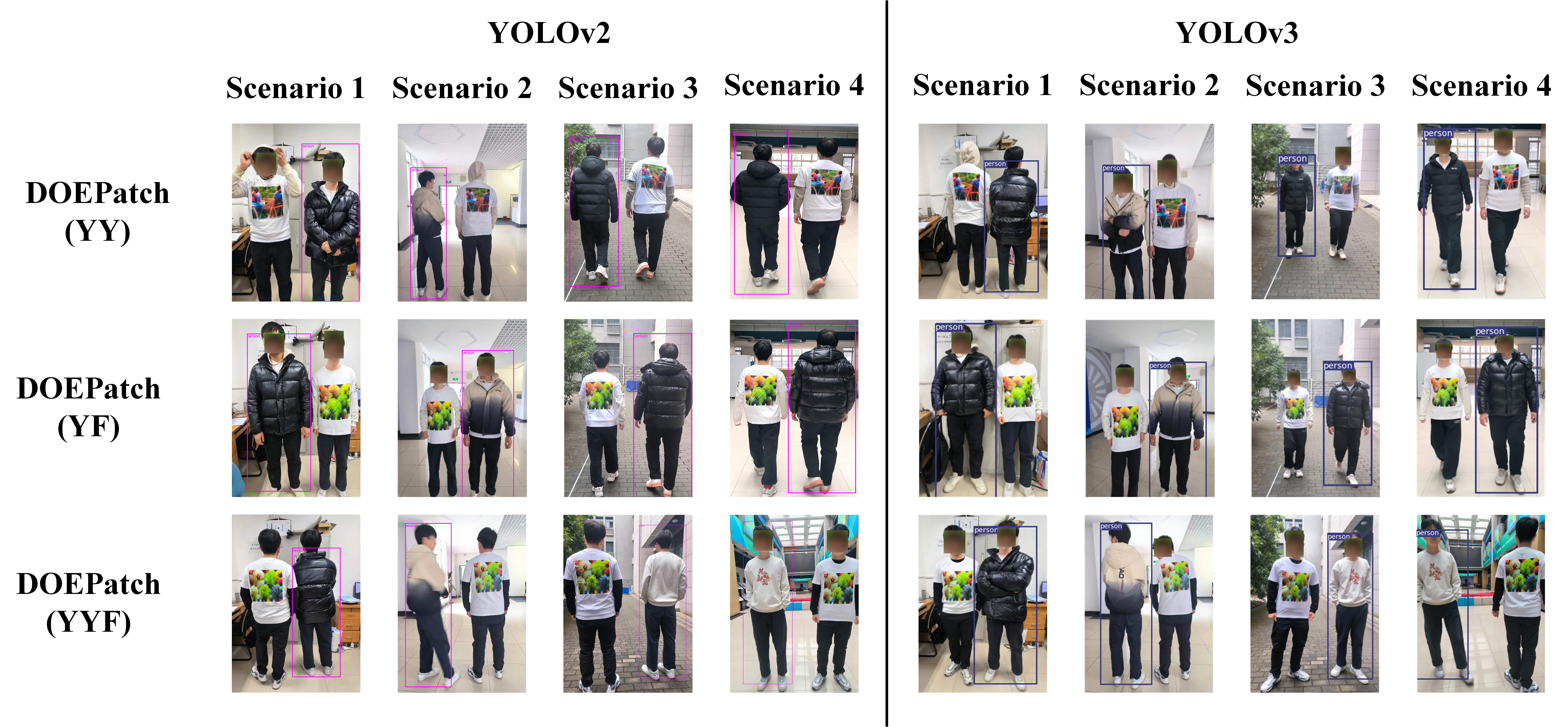}
\caption{In the detection results in the four scenarios, the adversarial patches printed on the front and back can affect the object detector recognition.}
\label{fig:7}
\end{figure*}

Partial prediction results as shown in Fig.\ref{fig:7} indicate that in different environments, both YOLOv2 and YOLOv3 can detect all individuals not wearing adversarial T-shirts in images. However, they cannot detect another person wearing adversarial T-shirts crafted using DOEPatch(YY), DOEPatch(YF), and DOEPatch(YYF). According to the results, the adversarial patches generated were found to be effective in manipulating the recognition capability of object detection models in the physical world. Specifically, the patches were able to enable individuals wearing adversarial T-shirts to evade detection by object detection models, while also exhibiting robust transferability across multiple object detection models. These findings highlight the potential vulnerability of real-world object detection models to adversarial attacks and emphasize the need for further research to enhance the robustness of such models against potential threats.

\subsection{Robustness Evaluations of the Adversarial Patch}
\label{sec:robust}
\begin{figure}[!t]
	\centering
	\includegraphics[scale=0.25]{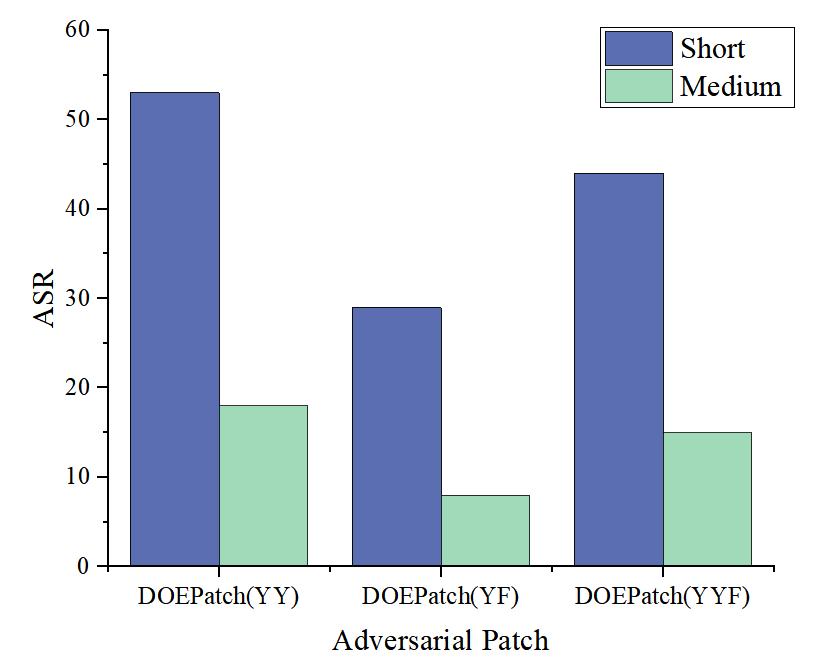}
	\caption{ASR test results of adversarial patches under different distances.}
	\label{fig:distance}
\end{figure}

To comprehensively assess the robustness of adversarial patches and their limitations in practical applications, we conducted tests based on two factors: distance and angle. Firstly, in each environment, we collected three datasets for each adversarial patch: short distance (2 to 4 meters), medium distance (5 to 7 meters), and long distance (8 to 10 meters), with 25 images per dataset. Therefore, each adversarial patch was validated on 100 images within each distance range. The experimental results, as shown in Fig.\ref{fig:distance}, were verified using YOLOv2. DOEPatch(YY) achieved ASR of 53\% at short distances, while DOEPatch(YYF) reached 44\%, indicating good attack effectiveness. However, with increasing distance, the attack effectiveness of adversarial patches decreased at medium distances. At long distances, we observed that all adversarial patches failed to impact the target detector's recognition. As the distance increased, the presentation of adversarial patches in the images became smaller and more blurred, leading to a decline in attack performance.

\begin{figure}[!t]
	\centering
	\includegraphics[scale=0.28]{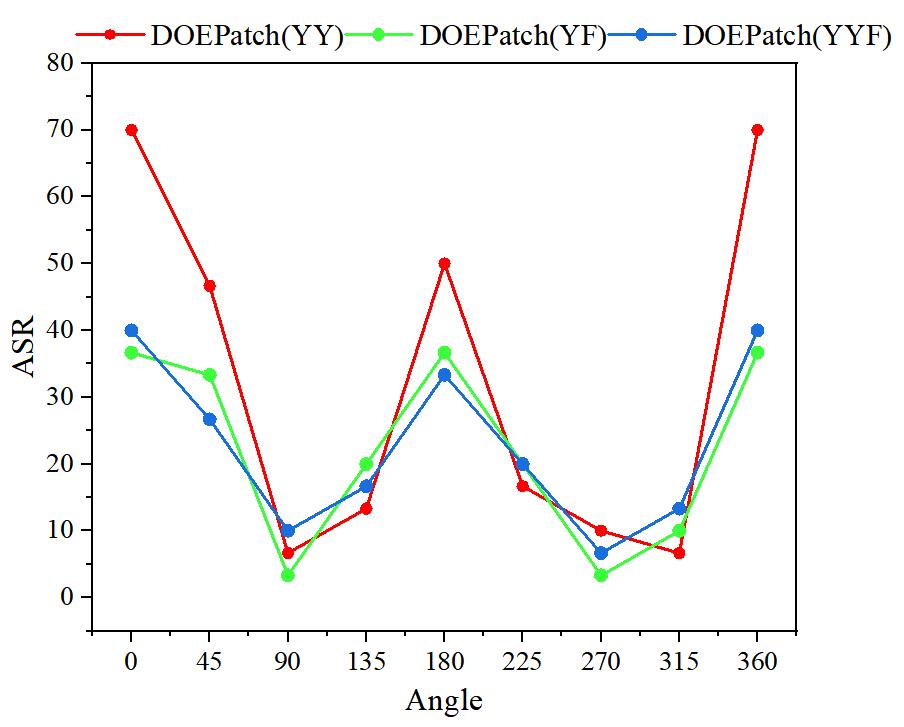}
	\caption{ASR test results of adversarial patches under different angles.}
	\label{fig:angle}
\end{figure}

For the angle transformation experiment, we collected datasets within the range of 0° to 360° at a distance of two meters from the target for each adversarial patch. The ASR test results on YOLOv2 are shown in Fig.\ref{fig:angle}. Each point of the marker was obtained by testing on 30 images. The adversarial patch DOEPatch(YY) had the best attack effect at 0° with an ASR of 70\%. As the angle changes, the attack effectiveness of all adversarial patches declines. However, after 90°, the ASR values begin to rise again, reaching a peak at 180°. Observing the trend of the curves in the graph, we find that adversarial patches exhibit good attack effectiveness only when directly facing the camera. Furthermore, frontal testing yields significantly better results than testing from the rear. In side-facing scenarios, due to the limitations of the viewing angle, most of the area covered by the adversarial patches is not captured, leading to a substantial decrease in attack effectiveness. Distance and viewing angle interference with adversarial patches are unavoidable, and what we need to do is to minimize the impact of these factors as much as possible. Increasing the area of the adversarial patches and addressing blurriness issues can help mitigate the effects of distance. Regarding the influence of viewing angles, considering a full coverage approach can reduce the impact of occlusion to some extent. These issues mentioned above are what we need to address in our next steps.

To evaluate the effectiveness of our method on a new model, we retrained the adversarial patch DOEPatch against the more advanced YOLOv5s.pt open-source model. The YOLOv5\cite{jocher10ultralytics} and YOLOv7\cite{wang2023yolov7} series of open-source models are more widely used in practical applications, with most real systems being based on these open-source models retrained to perform tasks. Therefore, we validated the attack effect of the adversarial patch on YOLOv5 and YOLOv7 models using the Inria test data, with the mAP testing results shown in Tab.\ref{lab:v5}. DOEPatch achieved a decrease of 39.2\% in mAP on the YOLOv5s.pt model. Even on the complex YOLOv7x.pt model, the mAP was reduced from 95.2\% to 59.4\%, indicating that our attack method exhibits good robustness across different models.

\begin{table*}[!ht]
	\centering
	\caption{The mAP testing results of adversarial patches in different models.}
	\begin{tabular}{c|c|c|c|c|c}
		\hline
		~ & YOLOv5s.pt & YOLOv5n.pt & YOLOv5x.pt & YOLOv7.pt & YOLOv7x.pt \\ \hline
		Clean & 0.934 & 0.921 & 0.959 & 0.957 & 0.952 \\ \hline
		DOEPatch & 0.542 & 0.627 & 0.613 & 0.651 & 0.594 \\ \hline
	\end{tabular}
	\label{lab:v5}
\end{table*}

\subsection{Interpretability of the Adversarial Patch}
\label{sec:4.4}

\begin{figure}[!t]
	\centering
	\includegraphics[scale=0.37]{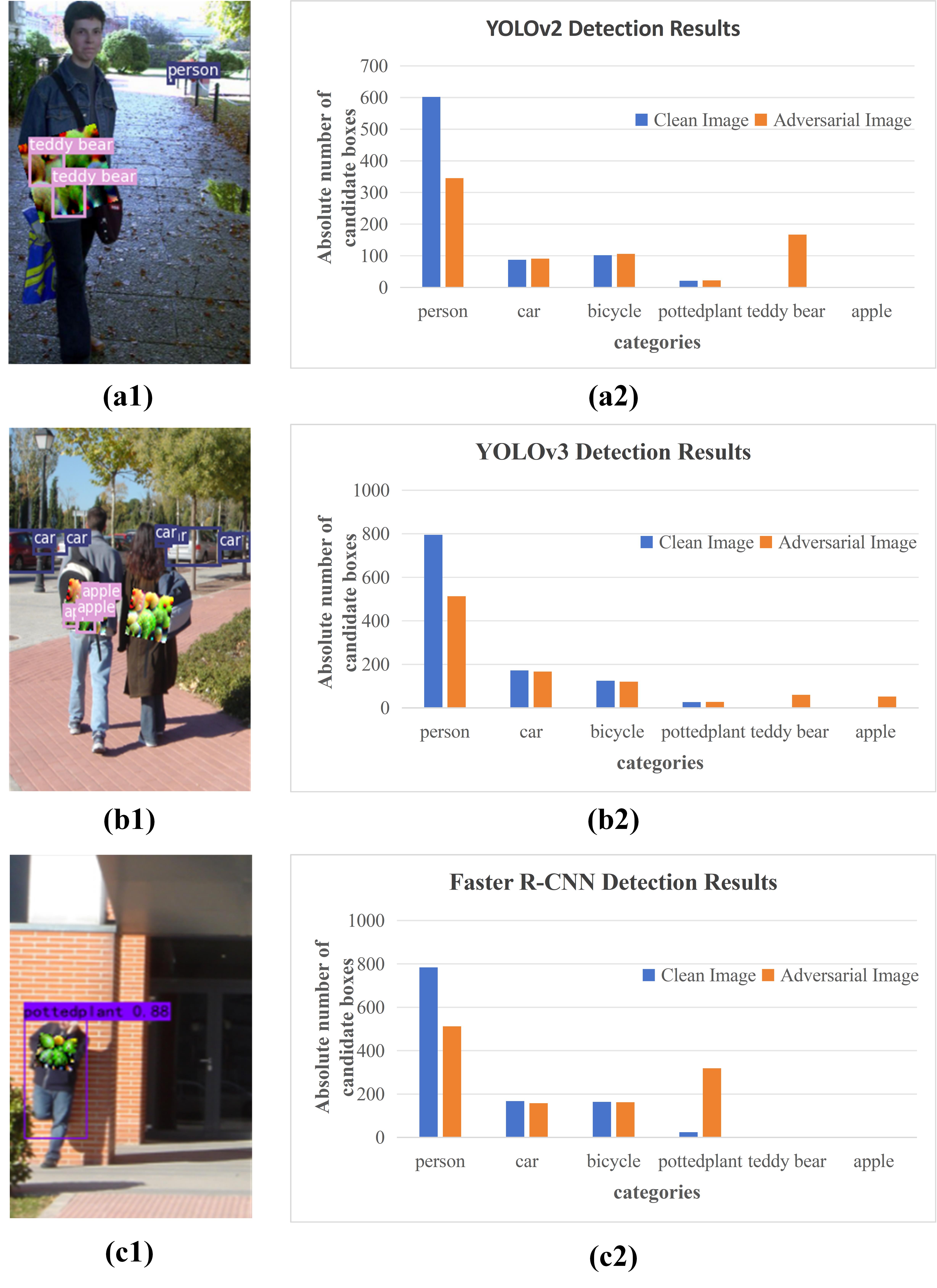}
	\caption{(a1), (b1), (c1) denote the results of misidentification of adversarial patches in object detection. (a2), (b2), (c2) denote the results of the change in the absolute number of predicted category candidate boxes.}
	\label{fig:8}
\end{figure}

Although most works have studied the interpretability of adversarial examples\cite{vellaichamy2022detectordetective}, there is no comprehensive explanation of the attack performance for adversarial patches.
Consequently, we offer a more comprehensive and insightful explanatory analysis of the phenomenon of how adversarial patches perturb target detection predictions, utilizing the energy function perspective and the visual interpretation tool of Grad-CAM\cite{selvaraju2017grad}. Adversarial patches generated by the energy function and the dynamically optimized ensemble model exhibit complex and distinct texture features in both color and feature distribution, resembling traditional adversarial patches. Fig.\ref{fig:8}(a1) shows the detection result of DOEPatch(YYF) in YOLOv2, misidentified as the ``teddy bear'' category. Fig.\ref{fig:8}(b1) depicts the detection result of DOEPatch(YF) in YOLOv3, misidentified as the ``apple'' category. Fig.\ref{fig:8}(c1) shows the detection result of EmPatch(F) in Faster R-CNN, misidentified as the ``potted plant'' category. Merely analyzing partial detection results is insufficient for uncovering the underlying mechanisms of adversarial patches. From the perspective of energy function analysis, the training process reduces the energy to impair the object detection capability of the target detector on the ``person'' category by optimizing the adversarial patch. However, it is challenging to visualize the energy change directly, and we describe the energy change process by comparing the number of selected category candidate boxes after the screening. As shown in the bar chart in Fig.\ref{fig:8}, the horizontal coordinates represent the predicted categories and the vertical coordinates represent the absolute number of candidate boxes of different categories output by the detector. Fig.\ref{fig:8}(a2) illustrates the changes in the candidate boxes' output of YOLOv2 upon adding DOEPatch(YYF) to the test samples. Fig.\ref{fig:8}(b2) shows the transformation of the candidate boxes' output of YOLOv3 after adding DOEPatch(YF) to the test samples, and Fig.\ref{fig:8}(c2) depicts the changes in the candidate boxes' output of Faster R-CNN after adding EmPatch(F). The blue bars represent the number of candidate boxes output by the detection network after predicting the original samples, while the orange bars represent the output after detecting the adversarial samples. Specifically, the number of candidate boxes representing the ``person'' category sharply decreases after adding the adversarial patch, while some other categories' quantities increase to a certain extent. By training adversarial patches with complicated features that enhance the energy of other categories, the energy of the person category is lowered to a certain extent. 

\begin{figure}[!t]
	\centering
	\includegraphics[scale=0.20]{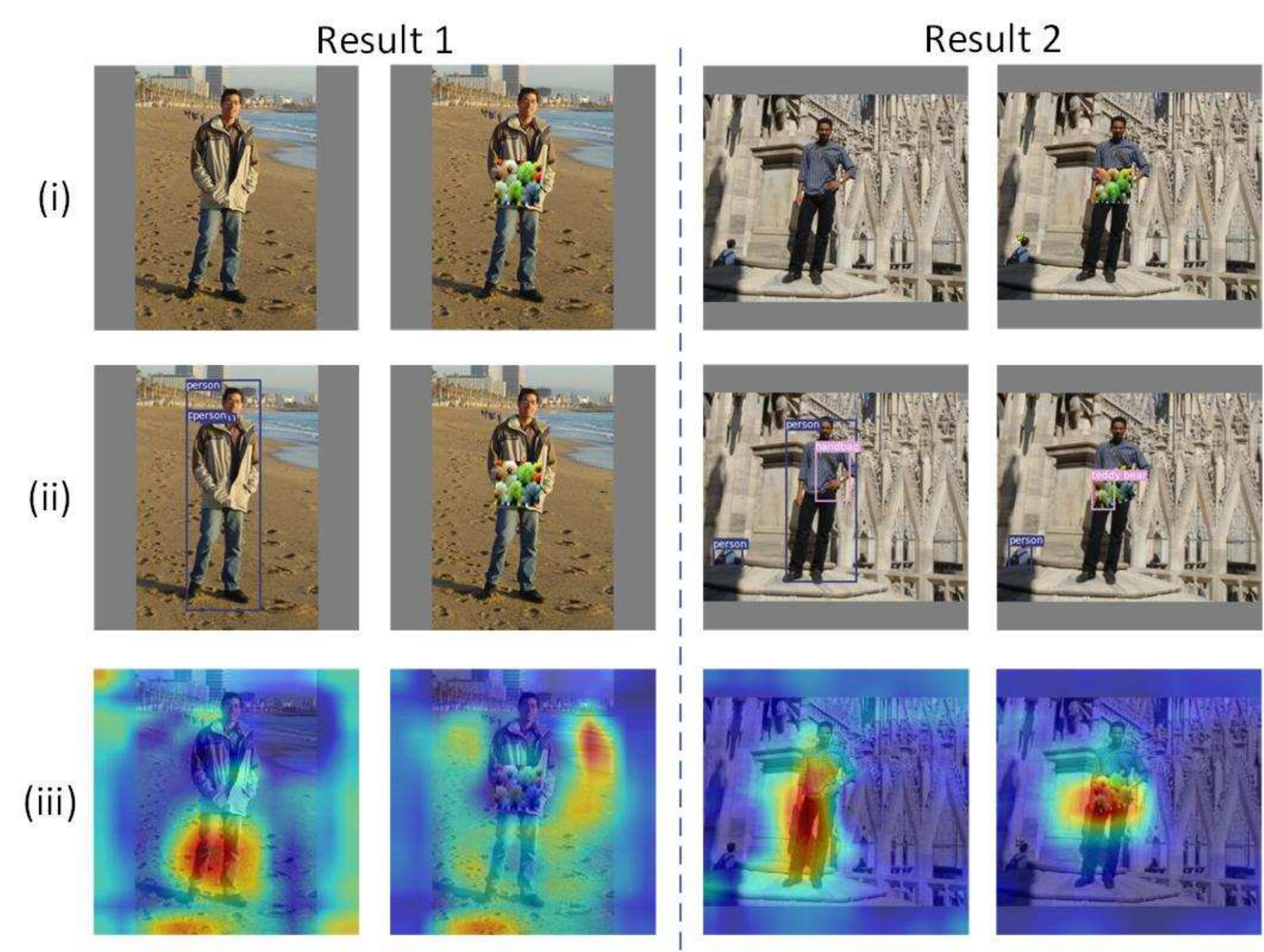}
	\caption{The attention distribution maps of the images before and after our attack. After adding the adversarial patches, the original distribution is disrupted.}
	\label{fig:9}
\end{figure}

Merely analyzing from the energy perspective is not comprehensive enough. In order to further analyze the effects of the adversarial attack, we used the pre-trained VGG16 model on ImageNet and combined it with the Grad-CAM algorithm to demonstrate the changes in the attention maps of the target images before and after the attack. Two main changes in the results were observed from the experiments conducted on the test dataset. As shown in Fig.\ref{fig:9}, the (i) row represents the original sample and the adversarial sample after adding the adversarial patch, the (ii) row represents the recognition results under the detector, and the (iii) row represents the detection results under Grad-CAM. As shown in (ii), in \textit{Result 1}, the person was not recognized as it was not attended to due to the attention being distracted by the adversarial patch. The feature distribution of the person was altered due to the presence of the adversarial patch. In \textit{Result 2}, the adversarial patch was misrecognized, and all the attention of the model was attracted to the patch, leading to a change in the model's decision. The adversarial patch has the ability to affect the distribution of the target features in the detection network, thereby interfering with the prediction results. 

To provide a more comprehensive analysis of the effectiveness of adversarial patches, we propose a quantitative metric called the Disruption Area Ratio $R_{d}$. This metric aims to measure the proportion of different areas within the bounding box region of the target object, as determined by the true label, in the Grad-CAM outputs of both clean images and adversarial samples. Specifically, we first generate Grad-CAM heatmaps for both clean images and adversarial samples. Then, based on the true label, we identify the bounding box region of the target object in the heatmaps. Subsequently, we compute the proportion of different regions between the two heatmaps, which serves to evaluate the impact of adversarial samples on the bounding box region of the target object. We validated this metric on the Inria Person test set, computing the average across 288 test images.
After adding the adversarial patch DOEPatch(YY), 60.92\% of the regions in the target object heat map distribution are different from the original image. With the addition of DOEPatch(YF) and DOEPatch(YYF), results of 68.27\% and 64.96\% were also obtained, respectively. This indicates that the feature distribution of the target object has been severely disturbed, resulting in an incorrect recognition by the target detector.

\subsection{Adversarial training Evaluations}
\label{sec:train}
Adversarial training entails introducing adversarial samples into the model, allowing it to learn features inherent in adversarial samples and thus bolstering the model's resilience against adversarial attacks. To validate the effectiveness of our generated adversarial samples in fortifying the reliability of object detection models, we devised adversarial training experiments for evaluation. Firstly, we randomly extracted 120 images from the Inria Person dataset for the training set, 30 images for the validation set, and 180 images for the testing set. Then, we added adversarial patches DOEPatch(YY), DOEPatch(YF), and DOEPatch(YYF) to the extracted images, resulting in adversarial sample training, validation, and testing sets. We added the generated adversarial sample training set to the original training sets of YOLOv3 and Faster R-CNN, and retrained the models. Therefore, the adversarial training mode we designed is as follows:
\begin{itemize}
	\item YOLO-YY: Add adversarial samples generated by DOEPatch(YY) to the original data, then retrain YOLOv3.
	\item YOLO-YF: Add adversarial samples generated by DOEPatch(YF) to the original data, then retrain YOLOv3.
	\item YOLO-YYF: Add adversarial samples generated by DOEPatch(YYF) to the original data, then retrain YOLOv3.
	\item FR-YY: Add adversarial samples generated by DOEPatch(YY) to the original data, then retrain Faster R-CNN.
	\item FR-YF: Add adversarial samples generated by DOEPatch(YF) to the original data, then retrain Faster R-CNN.
	\item FR-YYF: Add adversarial samples generated by DOEPatch(YYF) to the original data, then retrain Faster R-CNN.
\end{itemize}

We saved the model every 10 epochs and tested it on the validation set. After training for 100 epochs, we selected the model with the best results for subsequent testing. Tab.\ref{lab:training} shows the AP test results for all adversarial patches, where $\Delta$ represents the change in detection models after adversarial training compared to the original models. We observed significant improvements in the models after adversarial training, particularly in the testing set with adversarial samples. The YOLO-YY model achieved a 64.67\% increase in AP on DOEPatch(YY), while the FR-YYF model demonstrated a 33.53\% improvement on DOEPatch(YYF). Then, after adversarial training, the detection models exhibit strong resistance to interference from test data that possess features similar to the input adversarial samples. The YOLO-YF model achieved improvements of 48.01\% and 40.21\% on DOEPatch(YF) and DOEPatch(YYF) respectively. Meanwhile, the FR-YF model demonstrated enhancements of 33.93\% and 34.09\% on DOEPatch(YF) and DOEPatch(YYF) respectively. Furthermore, adversarial training not only enhances the model's resistance to interference but also improves its performance on clean samples compared to the original model. YOLO-YF exhibited a 3.93\% improvement on clean samples compared to the original model. Additionally, adversarial training does not necessarily grant the model stronger resistance to all adversarial patches. In the case of UpcPatch, the AP decreased in all tests involving YOLOv3 models. UpcPatch already yields high AP results when tested on the original model, making further enhancements challenging. Moreover, it possesses significant differences from the features of input adversarial samples, causing the model to lean towards recognizing the features of the input adversarial samples during training. In conclusion, although adversarial training may not render the model completely immune to adversarial patches, it can greatly enhance the robustness of detection models.

According to the experimental results of adversarial training, while training the detection model using adversarial samples as input data effectively improves performance, it still exhibits instability when faced with other unknown attacks. Some of the existing work on improving the adversarial training strategy based on adversarial samples to minimize the distance between clean and perturbed images, such as stylized adversarial training(SAT)\cite{naseer2022stylized}, attribute-guided adversarial training\cite{gokhale2021attribute}. Furthermore, the primary reason adversarial patches impact object detectors is their intricate surface texture and color distribution, which significantly influences the detector's ability to extract image features. Therefore, some preprocessing operations can be added to the input data to reduce the effect of the antagonistic samples, such as adding a denoising module\cite{shao2022open,ho2022disco} to filter the antagonistic noise. On the other hand, the vulnerability of the model itself is also an issue that cannot be ignored. The distillation-based\cite{papernot2016distillation,hinton2015distilling} approach compresses and transfers the knowledge from a complex large-scale model or an integrated model consisting of several models to a small model, which largely improves the stability of the small model. Hence, adversarial attack methods provide a new perspective for assessing the efficacy of models, emphasizing the significance of a model's robustness against unknown variations. Our approach enables a deeper investigation into the vulnerabilities of object detection models, offering novel directions for enhancing the security of object detection models.

\begin{table*}[!ht]
	\caption{The AP test results of detection models after adversarial training on adversarial patches are presented. The left axis lists all adversarial patches, while the top axis lists the detection models. $\Delta$ indicates the change in AP compared to the original models after adversarial training.}
	\centering
	\begin{tabular}{c|c|c|c|c|c|c|c|c}
		\hline
		~ & YOLOv3 & YOLO-YY($\Delta$) & YOLO-YF($\Delta$) & YOLO-YYF($\Delta$) & FR & FR-YY($\Delta$) & FR-YF($\Delta$) & FR-YYF($\Delta$) \\ \hline
		Clean & 89.25 & 92.18(2.93) & 93.18(3.93) & 90.41(1.16) & 92.77 & 92.85(0.08) & 93.32(0.55) & 93.36(0.59) \\ \hline
		AdvPatch & 71.21 & 87.71(16.5) & 88.68(17.47) & 88.25(17.04) & 70.3 & 85.34(15.04) & 82.02(11.72) & 83.26(12.96) \\ \hline
		TCEGA & 55.63 & 66.25(10.62) & 88.37(32.74) & 89.34(33.71) & 74.65 & 83.65(9.00) & 90.36(15.71) & 89.11(14.46) \\ \hline
		NaPatch & 76.89 & 76.14(-0.75) & 78.48(1.59) & 78.97(2.08) & 88.75 & 90.54(1.79) & 90.46(1.71) & 90.63(1.88) \\ \hline
		EmPatch(Y1) & 61.45 & 89.98(28.53) & 75.13(13.68) & 77.46(16.01) & 64.38 & 87.77(23.39) & 72.74(8.36) & 79.67(15.29) \\ \hline
		EmPatch(Y2) & 30.87 & 62.75(31.88) & 83.24(52.37) & 81.24(50.37) & 90.37 & 90.30(-0.07) & 93.66(3.29) & 90.77(0.40) \\ \hline
		DOEPatch(YY) & 26.17 & 87.84(61.67) & 59.47(33.3) & 63.35(37.18) & 64.82 & 87.50(22.68) & 75.14(10.32) & 80.77(15.95) \\ \hline
		EmPatch(F) & 83.77 & 81.98(-1.79) & 86.91(3.14) & 87.91(4.14) & 48.92 & 58.47(9.55) & 80.82(31.90) & 77.96(29.04) \\ \hline
		UpcPatch & 86.22 & 82.91(-3.31) & 76.48(-9.74) & 78.09(-8.13) & 90.35 & 89.14(-1.21) & 91.08(0.73) & 92.75(2.40) \\ \hline
		DOEPatch(YF) & 42.25 & 66.51(21.26) & 90.26(48.01) & 88.92(46.67) & 47.17 & 60.61(13.44) & 81.10(33.93) & 78.85(31.68) \\ \hline
		DOEPatch(YYF) & 47.86 & 76.82(28.96) & 88.07(40.21) & 89.01(41.15) & 48.49 & 66.55(18.06) & 82.58(34.09) & 82.02(33.53) \\ \hline
	\end{tabular}
	\label{lab:training}
\end{table*}

\subsection{Ablation Studies}
\label{sec:4.5}
To validate the effectiveness of the proposed dynamically optimized ensemble model, we conducted the following two experiments compared to the Average Ensemble model. 
We designed ensemble models utilizing the one-stage algorithms YOLOv2 and YOLOv3 and obtained respective adversarial patches, DOEPatch(YY) and AEPatch(YY). 
Additionally, we conducted experiments targeting Faster R-CNN and YOLOv3, which yielded novel adversarial patches, AEPatch(YF) and DOEPatch(YF). The weight parameters, denoted by ${{\text{w}}_0}$ and ${{\text{w}}_1}$, represent the weights before the detection model, which are constantly optimized during training in the dynamically optimized ensemble model. 
We manually adjusted the weight parameters of the average ensemble model and conducted five sets of experiments on AEPatch(YF), marked with numerical identifiers, as shown in Tab.\ref{lab:2}.
We evaluated the changes in AP across different detection models, and the detection results are presented in Tab.\ref{lab:3}. 
\begin{table}[!ht]
    \caption{AEPatch under different settings of ${{\text{w}}_0}$ and ${{\text{w}}_1}$.}
    \centering
    \begin{tabular}{c|c|c}
    \hline
        ~ & ${{\text{w}}_0}$ & ${{\text{w}}_1}$ \\ \hline
        AEPatch-1(YF) & 0.5 & 0.5 \\ \hline
        AEPatch-2(YF) & 0.2 & 0.8 \\ \hline
        AEPatch-3(YF) & 0.15 & 0.85 \\ \hline
        AEPatch-4(YF) & 0.1 & 0.9 \\ \hline
        AEPatch-5(YF) & 0.05 & 0.95 \\ \hline
    \end{tabular}
    \label{lab:2}
\end{table}
\begin{table}[!ht]
    \caption{The results of the comparative experiments between the two sets.}
    \centering
    \begin{tabular}{c|c|c|c}
    \hline
        ~ & YOLOv2 & YOLOv3 & Faster R-CNN  \\ \hline
        DOEPatch(YY) & 13.19\% & 29.20\% & 60.91\%  \\ \hline
        AEPatch(YY) & 15.62\% & 36.27\% & 62.44\%  \\ \hline
        DOEPatch(YF) & 59.64\% & 45.09\% & 47.10\%  \\ \hline
        AEPatch-1(YF) & 78.41\% & 82.66\% & 50.29\%  \\ \hline
        AEPatch-2(YF) & 76.20\% & 82.71\% & 47.83\%  \\ \hline
        AEPatch-3(YF) & 78.01\% & 82.06\% & 54.83\%  \\ \hline
        AEPatch-4(YF) & 72.39\% & 66.08\% & 68.85\%  \\ \hline
        AEPatch-5(YF) & 69.88\% & 40.25\% & 76.45\%  \\ \hline
    \end{tabular}
    \label{lab:3}
\end{table}

\begin{figure*}[!t]
\centering
\includegraphics[scale=0.3]{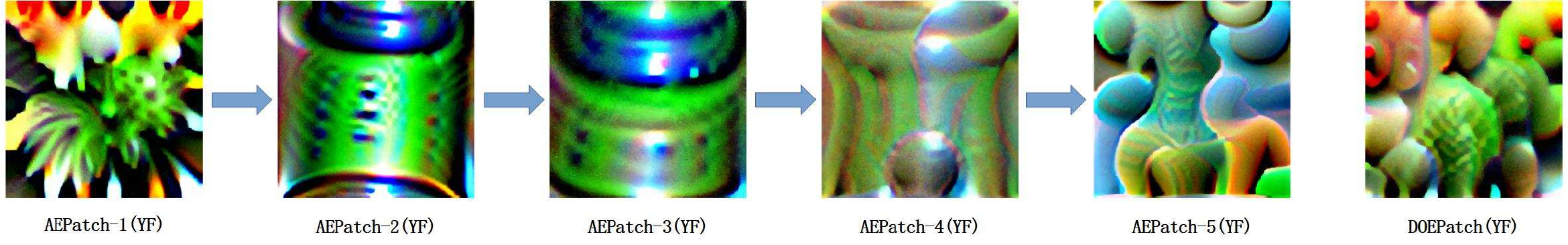}
\caption{The results of AEPatch(YF) after changes to its parameters.}
\label{fig:10}
\end{figure*}

In the first comparative experiment, both ensemble models achieved effective attack results, with DOEPatch(YY) exhibiting slightly better detection results than AEPatch(YY). In the second set of experiments, due to the more complex framework structure of the second-order detection algorithm, it was difficult for the average ensemble model to generate effective adversarial patches for both YOLOv3 and Faster R-CNN. Following manual parameter tuning across five sets, the training results revealed that AEPatch-4(YF) exhibited non-convergent behavior while overfitting was observed in the remaining patches. As shown in Fig.\ref{fig:10},  AEPatch-1(YF) exhibits a feature distribution that closely resembles the adversarial patch, EmPatch(F), trained solely on Faster R-CNN, whereas AEPatch-5(YF) displays a feature distribution more akin to that of the adversarial patch, EmPatch(Y2), trained independently on YOLOv3. Alterations to weight parameters result in shifts in the feature distributions of the adversarial patches. Achieving a balance point between the two distinct feature distributions is challenging through manual parameter tuning. However, DOEPatch(YF), which is generated by dynamic ensembling, addresses this issue and achieves favorable attack performance on both detection models. Therefore, the proposed algorithm addresses common issues of non-convergence and overfitting in ensemble model design, eliminates the complicated process of manual parameter tuning, and generates adversarial patches that can attack multiple detection models simultaneously. 

\section{Conclusion}
\label{sec:5}
In this paper, we propose a universal adversarial patch generation method based on dynamically optimized ensemble models, which can effectively attack multiple mainstream object detection models. Specifically, we design a transformation function for the adversarial patch that simulates the changes in patterns on clothing under real-world conditions. Then, we propose an adversarial patch generation algorithm for object detection models based on energy models. Finally, to enable the transferability of the adversarial patch across multiple detection models, we incorporate the concept of adversarial training and construct a dynamically optimized ensemble model. Through extensive comparative experiments, we demonstrate that our designed adversarial patch exhibits stronger attack performance and better transferability, with high attack effectiveness observed in physical-world scenarios. As the study of large-scale models continues to grow in popularity, it is critical to investigate the potential for a universal adversarial patch that can effectively target all mainstream object detection models. Our proposed algorithm provides new insights for future research on universal adversarial patches. 

\section*{Acknowledgment}
This work is funded by the National Natural Science Foundation of China (No.62103330, 62233014). 

\bibliographystyle{IEEEtran}
\bibliography{Ref}

\newpage

\vspace{11pt}

\vfill

\end{document}